\documentclass{article}

\usepackage{arxiv}
\usepackage[utf8]{inputenc} %
\usepackage[T1]{fontenc}    %
\usepackage{hyperref}       %
\usepackage{url}            %
\usepackage{booktabs}       %
\usepackage{amsmath}       %
\usepackage{amsfonts}       %
\usepackage{nicefrac}       %
\usepackage{microtype}      %
\usepackage{cleveref}       %
\usepackage{natbib}
\usepackage{doi}
\usepackage{balance}
\usepackage{xcolor}

\usepackage{mathtools}
\usepackage{algorithm}
\usepackage{algpseudocode}
\usepackage{wrapfig}
\usepackage{subcaption}
\usepackage{xspace}

\def\AAAIOURVERSION{1}  
\newcommand{\SD}{SD\xspace}  
\newcommand{\SDtext}{stable diffusion\xspace}
\newcommand{\SDText}{Stable diffusion\xspace}
\newcommand{\directed}{directed\xspace}  
\newcommand{\SSatK}{SS@k\xspace}
\newcommand{\SSat}{SS@\xspace}
\newcommand{\OneObject}{One Object\xspace}
\newcommand{\TwoObjects}{Two Objects\xspace}

\newcommand*{\ARXIVMERGEDAPPENDIX}{yes}%

\newcommand{\ifarxiv}[2]{#1}{}
\newcommand\AAAIedit[2]{{#1\xspace}}{}
\definecolor{purple}{rgb}{0.3, 0, 1.0}
\definecolor{darkred}{rgb}{0.6, 0.1, 0.1}

\newcommand{\checkthis}[1]{\textcolor{red}{#1}}

\newcommand{\FIXLATER}[1]{}  
\newcommand{\LONGVERSION}[1]{}  

\newcommand{\SecRef}[1]{{Sec.~\ref{#1}}}   
\newcommand{\FigRef}[1]{{Fig.~\ref{#1}}}   
\newcommand{\FigsRef}[1]{{Figs.~\ref{#1}}} 
\newcommand{\AlgoRef}[1]{{Algo.~\ref{#1}}}   
\newcommand{\EqRef}[1]{{Eq.~\ref{#1}}}
\newcommand{\link}[1]{\href{#1}{[link]}}
\newcommand{\Fig}{Fig.\ }

\newcommand{\Eq}{Eq.\ }

\renewcommand{\aa}{\mathbf{a}}
\newcommand{\bb}{\mathbf{b}}

\newcommand{\xx}{\mathbf{x}}

\newcommand{\zz}{\mathbf{z}}

\newcommand{\bAA}{\mathbf{A}}

\newcommand{\CC}{\mathbf{C}}
\newcommand{\DD}{\mathbf{D}}

\newcommand{\KK}{\mathbf{K}}

\newcommand{\MM}{\mathbf{M}}

\newcommand{\QQ}{\mathbf{Q}}

\newcommand{\VV}{\mathbf{V}}
\newcommand{\WW}{\mathbf{W}}

\newcommand{\BBB}{\mathcal{B}}

\newcommand{\DDD}{\mathcal{D}}

\newcommand{\III}{\mathcal{I}}

\newcommand{\LLL}{\mathcal{L}}

\newcommand{\PPP}{\mathcal{P}}

\newcommand{\RRR}{\mathcal{R}}

\newcommand{\TTT}{\mathcal{T}}

\newcommand{\Reals}{\mathbb{R}}	

\usepackage{amsmath}

\DeclareMathOperator*{\argmin}{arg\,min}

\AtBeginDocument{%
  }

\colorlet{texcscolor}{blue!50!black}
\colorlet{texemcolor}{red!70!black}
\colorlet{texpreamble}{red!70!black}
\colorlet{codebackground}{black!25!white!25}

\begin{document}

\title{Directed Diffusion: Direct Control of Object Placement through Attention Guidance}

\author{
    \href{https://orcid.org/0000-0002-9499-2623}{Wan-Duo Kurt Ma}\\
    Victoria University of Wellington\\
    \texttt{mawand@ecs.vuw.ac.nz}\\
\And
    \href{https://orcid.org/0000-0002-6835-7263}{J.P.~Lewis}\\  %
    Google Research\thanks{Current affiliation: NVIDIA Research}\\
    \texttt{noisebrain@gmail.com}\\
\And
    Avisek Lahiri\\
    Google Research\\
    \texttt{avisek@google.com}\\
\And
    Thomas Leung\\
    Google Research\\
    \texttt{leungt@google.com}
\And
    W. Bastiaan Kleijn\\
    Victoria University of Wellington and Google\\
    bastiaan.kleijn@vuw.ac.nz
}

\maketitle

\keywords{denoising diffusion \and text-to-image generative models \and artist guidance \and storytelling}
\vspace{0.5in}

\begin{abstract}
Text-guided diffusion models such as DALL$\cdot$E~2, Imagen, eDiff-I, and Stable Diffusion
are able to generate an effectively endless variety of images given only a short text prompt describing the desired image content. In many cases the images are of very high quality. However, these models often struggle to compose scenes containing several key objects such as characters in specified positional relationships.
The missing capability to ``direct'' the placement of characters and objects both within and across images is crucial in storytelling,
as recognized in the literature on film and animation theory.
In this work, we take a particularly straightforward approach to providing the needed direction.
Drawing on the observation that the cross-attention maps for prompt words reflect the spatial layout of objects denoted by those words, we introduce an optimization objective that produces ``activation'' at desired positions in these cross-attention maps. The resulting approach is a step toward generalizing the applicability of text-guided diffusion models beyond single images to collections of related images, as in storybooks.
Directed Diffusion provides easy high-level positional control over multiple objects,
while making use of an existing pre-trained model and maintaining a coherent blend between the positioned objects and the background. Moreover, it requires only a few lines to implement.\footnote{Our project page: \href{https://hohonu-vicml.github.io/DirectedDiffusion.Page}{https://hohonu-vicml.github.io/DirectedDiffusion.Page}}



\end{abstract}


\section{Introduction}

Text-to-image models such as DALL$\cdot$E 2 \cite{DALLE2}, Imagen \cite{IMAGEN}, eDiff-I \cite{ediffI}, and others have revolutionized image generation. Platforms such as Stable Diffusion \cite{huggingface} and similar systems have democratized this capability, as well as presenting new ethical challenges.

The forementioned systems generate arbitrary images simply by typing a "prompt" or description of the desired image. It is not always highlighted, however, that practical experience and experimentation are often needed if the user has a particular result in mind.
Text-to-image diffusion methods often fail to produce desired results, requiring repeated trial-and-error experiments with ``prompt engineering'' including negative prompts, different random seeds, and hyperparameters including the classifier-free guidance scale, number of denoising steps, and scheduler \cite{latentguide}.
This is particularly true for complex prompts involving descriptions of several objects.
For example, in Stable Diffusion a prompt such as \emph{``a bird flying over a house''} fails to generate the house with some seeds,
and in other cases renders both the bird and house but without the ``on'' relationship.
The time required for experimentation is made worse if optimization or fine-tuning computation is required, especially on a per-image basis.
These difficulties have led to the creation of ``Prompt \cite{promptmarketplaces}'' where expert users share and sell successful settings.

\begin{figure}[t]
  \centering
    \captionsetup{skip=3pt}
    \includegraphics[width=1.\linewidth]{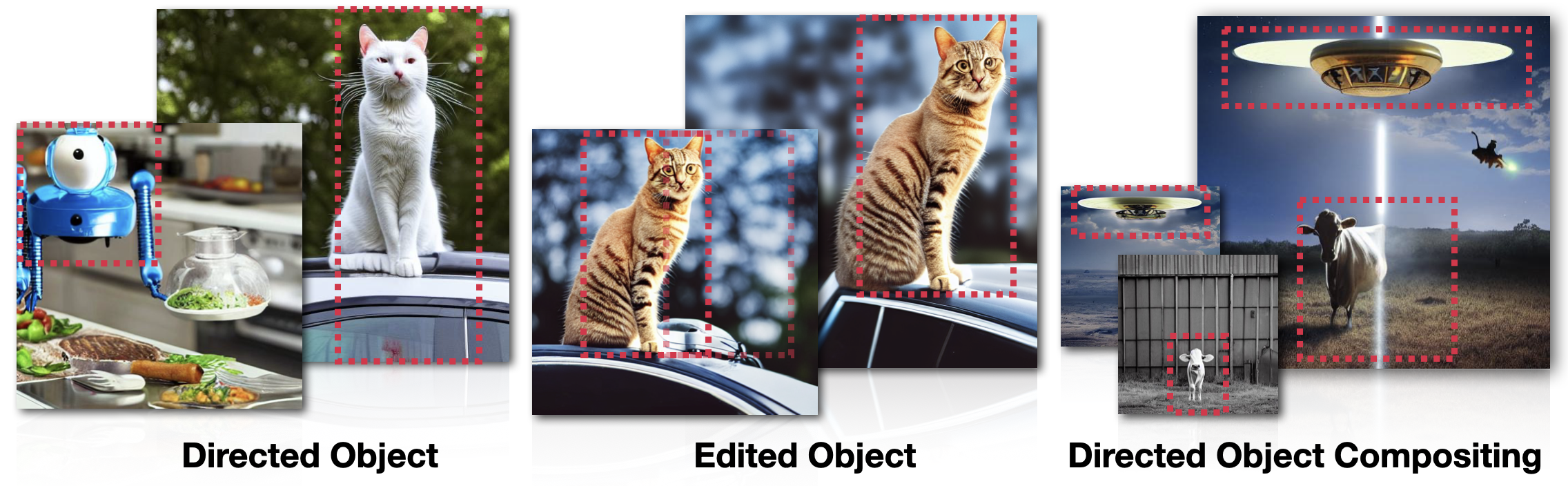}
    \caption{Directed Diffusion (DD) augments denoising diffusion text-to-image generation by allowing the position of specified objects to be controlled with user-specified bounding boxes (highlighted in red). (Left) DD generates specified objects (insect robot, cat) placed according to the given bounding boxes. We can move a synthesized object (middle), and place multiple objects in desired locations (right). All the directed objects show the appropriate ``contextual'' interaction (e.g.~shadows) with the background.
    }
   \label{fig:teaser}
\end{figure}

\ifdefined\AAAISHORTEN
The experimentation becomes prohibitive when the goal is to use the images for \emph{storytelling}, which involves
established principles for positioning characters relative to each other and the (virtual) camera \cite{FilmGrammar,IllusionOfLife}. Text prompts describing the content of an image do not indicate \emph{where} objects should be placed, and indicating desired positions in the prompt usually fails. As a consequence, extensive and tedious repeated trials are needed to obtain an image where the desired objects exist and their generated positions are acceptable.

\else

The experimentation becomes prohibitive when the goal is to use the images for \emph{storytelling},
since text-to-image methods offer no control over the required positioning of characters and objects. Prompts describing the content of an image do not indicate \emph{where} objects should be placed, and indicating desired positions in the prompt usually fails (\FigsRef{fig:teaser}, \ref{img:sd-prob}). This is because the training data of text-image pairs is gathered from public sources, and people rarely annotate the location of objects in an image
(\emph{``A photo of grandmother and her cat. Grandmother is in the center of the picture, and the cat is to her left.'')}
because it is generally obvious to the viewer. 
As a consequence, extensive and tedious repeated trials are needed to obtain an image where the desired objects exist and their randomly generated positions are acceptable.

A story generally involves a character interacting with the environment or with other characters. Conveying a story with images requires creating not just images with the desired semantic content, but images with objects in suitable relative \emph{positions} ("The thief looked back at the princess before hurrying toward the door").
The subject of "Film Grammar" \cite{FilmGrammar} seeks to codify established principles for positioning characters relative to each other and the camera. Film Language guides include principles such as
\emph{"To make things look natural, put lines, edges or faces about a third of the way across, up or down the picture ‘frame’. To make them look formal, put them in the middle; and to make things seem uncomfortable, make the shot unbalanced or put it at on a slant"} \cite{BeginnerFilmLanguage}.
The staging principle in traditional animation \cite{IllusionOfLife} similarly recognizes the importance of controlling the placement of characters and objects.
\fi

Research is addressing some limitations of text-guided diffusion methods, including methods that define new text tokens to denote specific and consistent character or object identities, 
provide mask-guided inpainting of particular regions, 
manipulate text guidance representations, 
and mitigate the common failure of guidance using CLIP-like models \cite{CLIP} to understand the target of attributes such as colors. 
Existing methods still generally struggle to synthesize \emph{multiple} objects with desired positional relationships (see the Experiment section).

Our work is a further step toward guiding text-based diffusion,
by introducing \emph{coarse positional control for objects} as needed for storytelling.
In this application only \emph{coarse} positional control is needed -- for example a director might instruct an actor to ``start from over there and walk toward the door'', rather than specifying the desired positions in floating point precision as is done in animation software packages. It is often necessary to directly control the position and interaction of \emph{two} objects, as in the case where a character is interacting with another character or object in some specified environment, however control over exact placement of background environment objects is rarely necessary or desirable.
We take inspiration from the observation that position is established early in the denoising process \cite{Magicmix} and from the fact that the cross-attention maps have a clear spatial interpretation 
(see \FigRef{img:sd-inter}).
Our general approach is to edit the cross-attention maps for particular words during the early denoising steps, so as to concentrate activation at the desired location of the object. We introduce an optimization objective that achieves this without disrupting the learned text-image association in the pretrained model and without requiring extensive code changes.

Our method is implemented using the Python \cite{diffusers} implementation of \SDtext (\SD)
and uses the available pre-trained model.

Our method makes the following contributions:
\begin{itemize}
\item \textbf{Storytelling.} Our method
is a step towards storytelling by providing consistent control over the positioning of multiple objects.
\item \textbf{Compositionality.} It provides a direct approach to ''compositionality'' by providing explicit positional control.
\item 
\textbf{Consistency.}

The positioned objects seamlessly and consistently fit in the environment, rather than appearing as a splice from another image with inconsistent interaction (shadows, lighting, etc.) This consistency is due to two factors. First, we use a simple bounding box to position and allow the denoising diffusion process to fill in the details, whereas specifying position with a pixel-space mask runs the risk that it may be inconsistent with the shape and position implicit in the early denoising results. Second, subsequent diffusion steps operate on the entire image and seamlessly produce consistent lighting,  etc. 
\item \textbf{Simplicity.}
Image editing methods for text-to-image models often require detailed masks, depth maps, or other precise guidance information. This information is not available when synthesizing images \emph{ab initio}, and it would be laborious to create. Our method allows the user to control the desired locations of objects simply by specifying approximate bounding boxes.

From a computational point of view, our method requires no training or fine tuning, and can be added to an existing text-driven diffusion model with cross-attention guidance with only a few lines of code.
It requires a simple optimization of a small weight vector $\aa \in \Reals^d, d < 77$, which does not significantly increase the overall synthesis time.
\end{itemize}
Storytelling also requires the ability to generate particular objects rather than generic instances
 (\emph{this} cat rather than ``any cat''), and our algorithm is complementary to methods  \cite{Textualinversion,Dreambooth} that address this. 
For clarity, \textbf{\emph{the examples in the paper do not make use of these other algorithms.}}

\section{Related Work}

Denoising diffusion models \cite{sohldickstein15,songScorematching19,hoDDPM20,DDIM} add Gaussian noise to the data in a number of steps and then train a model to incrementally remove this noise.
Aspects of the mathematics are presented in most papers and good tutorials are available \cite{blogLillog}, so we will simply mention several high-level intuitions. The end-result of the forward process is an image effectively consisting of independent normal distributed pixels, which is easy to sample. The backward process denoises random samples from this distribution resulting in novel images (or other data).
The mathematical derivation  \cite{sohldickstein15,hoDDPM20} somewhat resembles a hierarchical version of a VAE \cite{kingmaVAE2013}, although with a fixed encoder and a ``latent'' space with the same dimensionality as the data. The fixed encoder provides an easy closed-form posterior for each step, allowing the overall loss to split into a sum over uncoupled terms for each denoising step, resulting in faster training. 
\cite{songScorematching19} introduced an alternate derivation building on score matching \cite{Hyvarinen}. From this perspective adding noise is equivalent to convolving the probability density of the noise with the data, thus blurring the data distribution and providing gradients toward the data manifold from distant random starting points. The denoising process in \cite{hoDDPM20,songScorematching19} is stochastic, with an interpretation as Langevin sampling \cite{songScorematching19}.  A deterministic variant \cite{DDIM} is widely used for image editing applications.

Text-to-image (T2I) models condition the image generation process on 
the text representation from joint text-image embedding models such as CLIP \cite{CLIP},
thereby providing the ability to synthesize images simply
by typing a phrase that describes the desired image.
While T2I models have employed GANs as well as autoregressive models and transformers \cite{StyleCLIP,PARTI,MUSE}, 
a number of recent successful approaches use diffusion models as the underlying image generation mechanism \cite{GLIDE,DALLE2,IMAGEN}. This choice is motivated both by the stable training of these models and their ability to learn diverse multi-subject datasets.
However, T2I systems using CLIP are unable to reliably synthesize multiple objects (see supplementary). Often some objects are missing, or attributes such as color are applied to the wrong object \cite{TrainingFreeStructuredDiffusion}.

\ifdefined\AAAISHORTEN
\else
Among these systems, \SDtext (\SD) \cite{RombachStablediffusion21} has released both code and trained weights \cite{huggingface} under permissive license terms, resulting in widespread \AAAIedit{adoption.}{adoption and a large ecosystem of related tools.} This approach runs denoising diffusion in the latent space of a carefully trained autoencoder, providing accelerated training and inference.
\SDText implements classifier-free guidance using a cross-attention scheme. A projection of the latent image representation from \SD's U-net is used as the query, with projections of the CLIP embedding of the prompt supplying the key and value. While the key and value are constant,
the query is changing across denoising steps, allowing it to iteratively extract needed information from the text representation \cite{Perceiver21}.

The literature on applications of diffusion models is difficult to fully survey, with new papers appearing each day. A number of works have noted that the capabilities of text-guided diffusion can be extended with relatively simple modifications. \AAAIedit{}{For the case of image editing, SDEdit \cite{SDEdit} runs a given image part way through the noising (forward) process and then denoises the result. This has an interpretation of ''projecting on the image manifold'', and allows crude sketches to be denoised into sophisticated pictures. However there is a trade-off in the extent of the noising/denoising -- using the full process ``forgets'' all knowledge of the input image and just produces an unrelated random sample, whereas denoising for too few steps stays close to the input guidance image and inherits any of its imperfections.}
Blended Diffusion \cite{BlendedLatent} uses a user-provided mask to blend a noised version of the background image with the CLIP-guided synthesized inpainted region at each step. The authors point out an analogy to classic frequency-selective pyramidal blending, with the low frequency features being merged earlier in the denoising process. MagicMix \cite{Magicmix} produces convincing novel images such as \emph{``an espresso machine in the shape of a dog''}. 
Their technique simply starts the denoising process guided by the text prompt corresponding to the desired overall shape (``dog'') and switches to the prompt describing the content (``espresso machine'') at some step in the denoising process.
This exploits the observation that the early denoising steps establish the overall position and shape of an object while the later steps fill in the ``semantic details'' (\FigRef{img:sd-inter}). 
Prompt-to-prompt \cite{PromptToPrompt} demonstrates that powerful text-driven image editing can be obtained by substituting or weighting the cross-attention maps corresponding to particular words in the prompt.
The paper also clearly illustrated the fact that the cross-attention maps have a spatial interpretation.
Structured diffusion guidance \cite{TrainingFreeStructuredDiffusion} addresses the attribute binding problem in which text-to-image models often fail to associate attributes such as colors with the correct objects.
Their approach parses the prompt to obtain noun phrases with their associated attributes, and then these additional text embeddings are combined with that for the original prompt in generating the cross-attention.
\fi  


Complementary to the T2I problem, other research e.g.~\cite{layout2im,L2Igan,L2Ixformer} has addressed the layout-to-image (L2I) problem of positioning objects in an image. However, these methods often use closed-set training supervision, i.e.~they are unable to position object categories that were not part of the training. 

Our goal is to help democratize storytelling by enabling high-level open-set, zero-shot placement of several objects
in T2I denoising-diffusion synthesis, \emph{while exploiting a pre-trained model} to avoid requiring  computational resources beyond those available to the typical user. 

To achieve this we note the following criteria:
\begin{itemize}
\item
While a number of methods e.g.~\cite{BlendedLatent,Spatext} use \emph{shape masks} (silhouettes) to guide object placement, we intentionally avoid the use of detailed shape masks, because non-artists are generally not able to produce plausible outlines of objects in perspective (see supplementary). In addition, it has been noted that some shape mask methods produce poor alignment of the synthesized object to the given shape \cite{ShapeGuided} and may show poor interactions (e.g.~shadows) between the object and background. Our approach makes use of a small optimization to guide the cross-attention maps to seamlessly place the directed object. We find that this gives better quality and is more reliable than approaches that directly edit the cross-attention maps (see Fig.~\ref{img:trailing} in the supplementary).

\item
We  desire the ability to control the placement of a ``hero'' character or object, as well as (possibly) the position of another object that the hero is interacting with, 
while including both physical interactions (e.g. shadows) and (to the extent possible) storytelling interactions (``the fox \emph{chased} the rabbit''), along with optional description of the environment (``in the forest'').
\cite{LiuCompositional22} provides control over multiple objects but requires training and does not give position control.
Other work (\cite{multidiffusion}) demonstrates large scene composition with multiple objects but does not demonstrate specified interactions between placed objects.
\item
\cite{multidiffusion} note the increasingly important distinction between methods that require costly training on curated datasets and those that control the generated content by manipulating the generation process of a pre-trained model. While training custom models or fine-tuning \cite{Spatext,GLIGEN} offers the most power and flexibility, we exclude these approaches since they often require computational resources and data outside the reach of typical users. For example \cite{controlnet} reports requiring 100s of hours of GPU time for fine-tuning \SD while \cite{GLIGEN} uses 16 V100 GPUs.
We base our approach on \SD since it is not proprietary.  
\end{itemize}

Recent papers and preprints \cite{GLIGEN,multidiffusion,ediffI,boxdiff} provide robust control over the placement of multiple objects.  \cite{ediffI} is a powerful trained-from-scratch system that demonstrates a paint-with-words interface in which multiple objects are guided by both a prompt and a coarse shape mask.  \cite{ShapeGuided} 
demonstrate high-quality editing of a single object driven by a combination of text and a detailed mask. They address the issue that the object shape arising from classifier-guided diffusion can conflict with the provided mask. 
Similar to our work, \cite{GLIGEN} argues that high-level box guidance has advantages over shape masks, however their approach requires fine-tuning. \cite{boxdiff} and \cite{multidiffusion} are most similar to our aims. Like ours, these systems make use of a pretrained T2I model with no fine-tuning required, and both involve a relatively lightweight placement optimization during synthesis, although the particular approaches differ. We only recently became aware of these concurrent developments and did not perform a detailed evaluation, however brief comparisons are provided in the results section and supplementary. 

 \cite{pan-autoregressivestory-22} introduces an alternate approach to synthesizing images for storytelling, using an autoregressive formulation of latent diffusion in which each new image is conditioned on previous captions and images. This approach produces good results, however the positions of subjects are not controlled, making it difficult to use position-based storytelling principles.


\section{Method}
\label{sec:method}

Our objective is to create a controllable synthetic image from a text-guided diffusion model without any training by manipulating the attention from cross-attention layers, and the predictive latent noise. We use the following notation: Bold capital letters (e.g., $\MM$) denote a matrix or a tensor, vectors are represented with bold lowercase letters (e.g., $\textbf{m}$), and lowercase letters (e.g., $m$) denote scalars.
Depending on the context, the superscript $i$ on a three-dimensional tensor (e.g., $\MM^{(i)}$) denotes a tensor slice (a matrix). In this section, this index specifies the slice of the cross-attention map associated with a particular token 
in the prompt. Similarly, $\MM^{(i:j)}$ denotes the 
stacked slices $i$ to $j$ of the tensor.

\begin{figure}[t]
    \centering
    \includegraphics[width=0.8\linewidth]{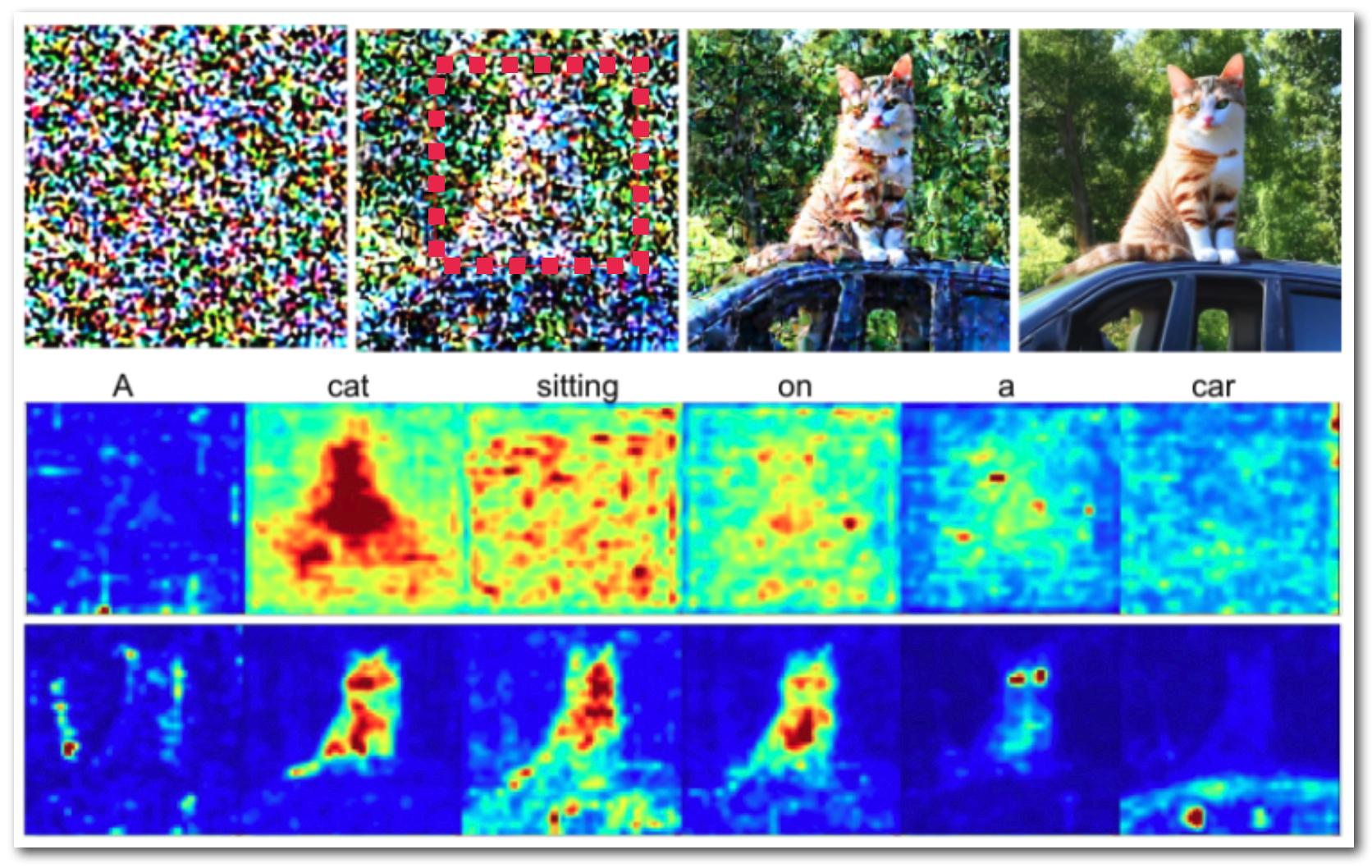}
    \caption{(Top, from left to right): The reverse \SD denoising process from the initial stage to the end of process. Note that the position of the cat is evident early in the process (red box), however the details that define it as a cat are not yet clear. (Bottom): The cross-attention maps associated with each word in the prompt, where the first and second row are from the first denoising step, and the final step, respectively. The cat's shape is clearly visible in the cross-attention maps for the word "cat" (and in the maps for "sitting" and "on" in the final step).}
    \label{img:sd-inter}
\end{figure}

The DD procedure controls the placement of objects corresponding to several groups of selected words in the prompt; we refer to these as \emph{\directed objects} and \emph{\directed prompt words} respectively. 
Our method is inspired by the intermediate result shown in \FigRef{img:sd-inter} (Top). As shown in this figure, the overall position and shape of a synthesized object  appears near the beginning of the denoising process, while the final denoising steps do not change this overall position but add details that make it identifiable as a particular object (cat). This observation has also been exploited in previous work such as \cite{Magicmix}.

An additional phenomenon can be found in the cross-attention maps, as shown in \FigRef{img:sd-inter} (bottom two rows).
The cross-attention map has a spatial interpretation, which is the ``correlation'' between locations in the image and the meaning of a particular word in the prompt. For instance, we can see the cat shape in the cross-attention map associated with the word ``cat''. DD utilizes this key observation to spatially guide the diffusion denoising process to satisfy the user's requirement.

\subsection{Pipeline}
\label{ss:pp}

Given the prompt $\PPP$ and the associated region information $\RRR$ indicating the directed objects,
DD synthesizes an image $\xx_0$ 
with appropriate ``contextual interaction'' between the objects and the remainder of the scene.
It uses a pre-trained Latent Diffusion Model (LDM) \cite{latentguide} no further fine-tuning. The region information $\RRR$ comprises a set of parameters $\RRR = \{\BBB, \III\}$, denoting the bounding boxes to position the directed objects, and the directed prompt word indices in the prompt. We will describe these in \SecRef{subsec:cam}.

\begin{figure*}[th]
    \captionsetup{skip=3pt}
    \centering
    \includegraphics[width=1.\textwidth,trim=0 0 0 7,clip]{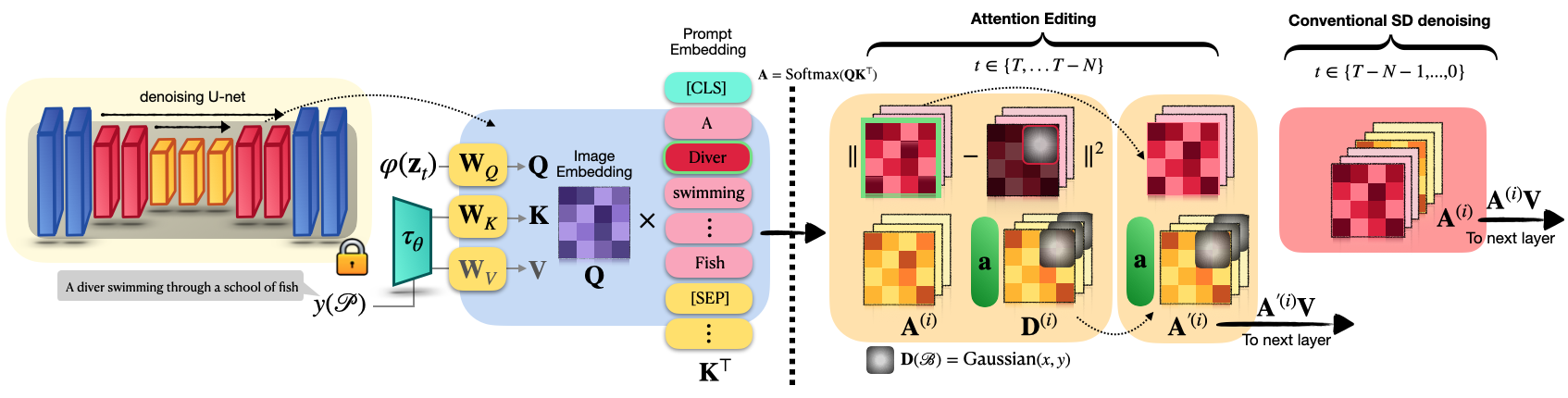}

\iftrue
 \caption{
Directed Diffusion (DD) pipeline overview: DD divides the denoising process into initial steps where \emph{Attention Editing} is performed,
followed by refinement steps using \emph{Conventional \SD Denoising}.
The goal of the \emph{Attention Editing} stage is to optimize the cross attention map for a directed word (red, outlined in green) to approximately match a target $\DD$ in which neural ``activation'' has been injected with a Gaussian fall-off inside a bounding box specified by the user. The optimization is performed by adjusting a vector ($\aa$, green) that re-weights the trailing attention maps. Note that the optimization objective involves sulccessive two time steps. Please see the text for details.}
\fi
\label{img:pipeline}
\end{figure*}

Following conventional notation, we define $\xx_t$ and $\zz_t$ as the synthesized SD image and the predicted latent noise at time step $t$, respectively, where $t \in \{T,...,0\}$ in the denoising fashion. The images $\mathbf{x}_t$ are reconstructions  obtained by feeding the latent $\mathbf{z}_t$ through the VAE decoder $\DDD(\cdot)$. The images $\mathbf{x}_T$ and  $\mathbf{x}_0$ correspond to the Gaussian noise latent $\zz_T$ and the final predicted latent $\zz_0$, respectively.

The principle of this work is based on the concept \emph{``first position the objects, then refine the results''.} This is reflected in the overall Directed Diffusion pipeline shown in \FigRef{img:pipeline} and also the applications detailed in \SecRef{s:app}.

\emph{Attention Editing}.
From a high-level perspective, this stage focuses on spatially editing the cross-attention map used for conditioning in \SDtext. It operates during the diffusion steps $t \in [T, T{-}N)$ that establish the object location, where $N$ is a hyperparameter determining the number of steps in this stage. We chose $N{=}10$ in most of our experiments. As described in \SecRef{subsec:cam}, this stage modifies the cross-attention map during the first $N$ denoising steps by  amplifying the region inside $\BBB$ while down-weighting the surrounding areas through optimization (the path to the yellow regions in \FigRef{img:pipeline}).

\emph{Conventional SD Denoising}. Following the attention editing stage, this stage runs the standard \SD process using classifier-free guidance \cite{hoClassifierfree21} over the remainder of the reverse diffusion denoising steps $t \in \{ T{-}N, \cdots, 0 \}$. Note that the only difference between the two stages is the cross-attention editing (the path to the red region in \FigRef{img:pipeline}).

\subsection{Cross-Attention Map Guidance}
\label{subsec:cam}

To achieve our goal, DD asks the user specific information about the ``direction'' of the object with $\RRR = \{\BBB, \III\}$ to guide the denoising SD process, where $\BBB$ and $\III$ denote the bounding box specification and the directed prompt words indices in $\PPP$, respectively. We will use the prompt \emph{``A bear watching a flying bird''} as our specific example throughout this section and the next section.

Specifically, a bounding box $\BBB = \bigl\{ (x, y) \,| \, b_{\text{left}} \times w \leq x \leq b_{\text{right}} \times w, \; b_{\text{top}} \times h \leq y \leq b_{\text{bottom}} \times h \bigl\}$ is the set of all pixel coordinates inside the bounding box of resolution $w \times h$. The bounding box with label $i$ guides the location of the subject indicated by the $i$th prompt word. The exact location and shape of the subject results from the interaction of the Gaussian activation window inside this box with the activation $\zz_i$ in the U-net, hence the object is not restricted in shape and may extend somewhat outside the box.
In our implementation, $\BBB$ is generated based on a tuple of four scalars representing the boundary of the bounding box, denoted as $\bb = (b_{\text{left}}, b_{\text{right}}, b_{\text{top}}, b_{\text{bottom}})$, where $b_{\text{left}}, b_{\text{right}}, b_{\text{top}}, b_{\text{bottom}} \in [0,1]$ describe the bounding box of the directed object position expressed as fractions of the image size, 

\newcommand{\NW}{|\mathcal{W}|}

\SD implements text-guided synthesis with classifier-free guidance using cross-attention
between the projected embedding $\QQ$ of the \SD latent $z_t$ and the projected embeddings $\KK$ of the $\NW$ potential words from the prompt.
The product $\QQ \KK^T \in  \Reals^{n^2 \times \NW}$ (following a softmax and normalization) can be interpreted as $\NW$ individual cross-attention maps $\bAA^{(i)}$ (see \FigRef{img:sd-inter}, bottom), each having a spatial interpretation of an $n \times n$ image.
 \SD uses CLIP \cite{CLIP} as its text encoder, which has has $\NW=77$, while $n=64$, matching the latent spatial dimension in the current $\SD$ implementation. The cross-attention maps $\bAA^{(i)}$ are divided into a subset
 of \emph{prompt maps}
  $\bAA^{(i)}, \, 1 \leq i \leq |\PPP|$ corresponding to words in the prompt, and the remainder $i \in \TTT$ where $\TTT \coloneqq \{|\PPP|{+}1,\cdots,\NW\}$
 that do not correspond to any words but are computed to allow constant-size matrix computations. We term the latter as \emph{trailing attention maps}.
We denote the prompt words for an object to be directed with an indices set $\III \subset \{ i | i \in \mathbb{N}, 1 \leq i \leq |\PPP| \}$.  For the bear example, $\III = \{1,2\}$ to denote ``A'' and ``bear'' in the prompt.

We wish to edit the cross attention maps corresponding to the ``directed'' words in the prompt so as to place the object at the bounding box. In one initial experiment, we used an optimization objective that encouraged the latent variables $\zz_t$ to have high energy in the desired region during the attention editing steps. This was not stable, which might be explained by the fact that it is potentially pushing $\zz_t$ outside the range of values encountered during training.

Thus, we wish to guide the directed cross-attention maps, but without disrupting the image-text relationship that is learned during training. One strategy is to use the network itself to project the solution near the ``trained manifold'', by expressing an objective at time $t$ in terms of coarser guidance made at time $t+1$ that is refined during the $t+1 \rightarrow t$ denoising step. Another observation is that we can separate the desired coarse guidance from the
directed cross attention maps by making
use of the trailing maps to express the coarse guidance, since these are also included in the $\QQ \KK^T$ product (see supplementary for further discussion).

Given $\BBB$, we generate a modified cross-attention map using two functions:
\[
\text{WeakenMask}(\BBB')_{xy} =
    \left\{\begin{matrix}
        c, \quad (x, y) \in \BBB' \\
        1., \quad \text{otherwise},
    \end{matrix}\right.
\]
\[
\text{StrengthenMask}(\BBB)_{xy} =
    \left\{\begin{matrix}
        f(x, y), \quad (x, y) \in \BBB \\
        0, \quad \text{otherwise},
    \end{matrix}\right.
\]
where $\BBB'$ is the complement of $\BBB$, and
$f(\cdot)$ denotes the function that ``injects attention'' to amplify the region $\BBB$. 
In our implementation, we use a Gaussian window of size $\sigma_x = b_w/2, \sigma_y = b_h/2$ to generate the corresponding weight, where $b_w = \text{ceil}((b_\text{right} - b_\text{left}) \times w), b_h = \text{ceil}((b_\text{top} - b_\text{bottom}) \times h)$
are the width and the height of $\BBB$.

The $\text{WeakenMask}(\cdot)$ and $\text{StrengthenMask}(\cdot)$ functions
``direct'' selected subjects from the prompt toward specific locations of the image in the \SD denoising process.
$\text{StrengthenMask}(\cdot)$ inserts higher activation into the provided bounding box with a Gaussian window, while $\text{WeakenMask}(\cdot)$ attenuates the region outside $\BBB$ by multiplying by $c < 1$.


The resulting edited maps $\DD^{(i)}$ are termed \emph{target} maps. They are not used directly in the cross-attention denoising guidance. Instead, as described next, $\DD^{(i)}$  $\, \forall i \in \III$ provide a target for the loss in \EqRef{eq:mainloss},
while $\DD^{(i)}$ $\forall i \in \mathcal{T}$ are un-weighted sources for the weighted trailing maps created in Algo.~\ref{algo:ddcae} line 12.

\iftrue


\begin{figure}[th]
\begin{algorithm}[H]
\caption{DD Cross-Attention Editing Algorithm}\label{euclid}
\label{algo:ddcae}
\begin{algorithmic}[1]
\State \textbf{Input:} 
diffusion model $\text{DM}(\cdot)$, 
prompt $\PPP$, 
directions $\RRR = \{\BBB,\III\}$, 
  timestep $t$
\State \textbf{Output:} Predicted conditional noise $\zz_{\text{cond}}$ at time $t$ from $\text{DM}(\cdot)$
\State \textbf{Parameters:} Gaussian weighting scalar $c_g$
\Procedure{DDCrossAttnEdit}{$\text{DM}(\cdot)$, $\PPP$, $\BBB$}
    \For{$l \in \text{layer}(\text{DM}( \zz_t,\PPP ))$}
            \If{$\text{type}(l) \in \text{CrossAttn}$}
                \State $\bAA = \text{Softmax}(\QQ_l(\zz_t) \cdot \KK_l(\PPP)^T)$
                \State $\WW \gets \text{WeakenMask}(\BBB')$
                \State $\mathbf{S} \gets \text{StrengthenMask}(\BBB)$
                \State $\DD^{(i)} \gets \bAA^{(i)} \odot \WW + \cdot \mathbf{S} \quad\forall i \in \TTT \cup \III$  
                \State $\aa^{\ast} \coloneqq \argmin_{\aa}\, \LLL_{\aa}$
                \State $\bAA'^{(|\PPP|+1:77)} \coloneqq \text{Diag}(\aa^{\ast}) \odot \DD^{(|\PPP|+1:77)}$
                \State $\zz_t \gets l(\zz_t, \bAA' \cdot \VV_{l}(\PPP))$
            \Else
                \State $\zz_t \gets l(\zz_t)$
            \EndIf
        
    \EndFor
    \State \textbf{return} $\zz_{\text{t}}$
\EndProcedure
\end{algorithmic}
\end{algorithm}
\end{figure}

\fi

\AlgoRef{algo:ddcae} is the core of DD. \SD implements denoising with a U-net architecture, where text guidance $\tau_\theta(y(\mathcal{P}))$ from the prompt is passed to layers of the U-net through cross-attention layers, with $y(\cdot)$ and $\tau_\theta(\cdot)$ denoting the tokenizer and text embedding, respectively. In each of these cross attention layers, DD clones and modifies  selected cross-attention maps $\bAA^{(i)}, \forall i \in \III \cup \TTT$ to form the target maps $\DD$ (line 10) for the optimization objective (line 11, \EqRef{eq:mainloss}).

Our optimization objective seeks to find the best weighed combination of the trailing maps at time $t$, by means of a learned weight vector  $\aa \in \Reals^{77-|\PPP|-1}$, such that the ``directed'' prompt maps $\bAA^{(i)} , i \in \III$  best match the corresponding target maps $\DD^{(i)}$  (with some abuse of notation):

\begin{equation}
    \LLL_{\aa_t} = \sum_{i} \| \bAA_{t-1}^{(i)}\left( \bAA_{t}^{(|\PPP|+1:77)} \cdot \text{Diag}(\aa_{t})\right) - \DD^{(i)}\|^2,
\label{eq:mainloss}
\end{equation}

$\forall t \in     \{T,...,T{-}N\}$, where the notation $\bAA_{t-1}^{(i)}\left(\cdot\right)$ indicates an (indirect) \emph{functional} dependence of the cross-attention map at time $t{-}1$ on the weighted sum of the trailing maps edited at time $t$.
After the weights $\aa^{\ast} = \argmin_\aa \LLL_{\aa_t}$ are obtained (line 11), they are used to produce re-weighted trailing maps $\bAA'$ (line 12),
which then influence the overall cross-attention conditioning at time $t$ (line 13).

To summarize the dependency flow,
 the trailing maps at time $t$ are edited,
and are reweighted by the optimized $\aa$.
These then influence the denoising at time $t$,
which in turn influences the cross attention map at time $t{-}1$.
This dependency structure involving expressing the loss at time $t{-}1$ is
necessary because at time $t$ the loss does not involve the weights $\aa$, resulting in a zero gradient. In addition we believe that influencing the desired cross-attention through an intermediate denoising step
helps keep the solution closer to states seen in training and thus removes the instability we found when directly optimizing for $\zz_t$. Please refer to the Appendix for further implementation details.



\section{Applications}
\label{s:app}
This section demonstrates two applications that can be utilized following the DD reconstruction process.  Both methods re-use information recorded from the DD processes.
In the descriptions, recall that in SD algorithms the latent $\zz_t$ ($64 \times 64$ in current implementations) has direct a spatial interpretation.

\subsection{Scene compositing}
\label{ss:sc}


Composing multiple objects is a challenging problem in image generation.
Complex prompts involving a relation between two objects \emph{(``bear watching bird'')} often fail in current T2I models.
While our DD pipeline supports the direction of multiple objects by means of multiple bounding boxes $\BBB$,
in practice, the results are unreliable when the number of bounding boxes is more than two. We resolve this problem by additional editing operations.

Specifically, we first use DD to individually generate multiple objects, and record the latent information $\zz^{(r)}_{t}$ of all steps for the directed objects $r \in \{1,\cdots,R\}$, where $R$ is the total number of directed objects. Then, inspired by \cite{LiuCompositional22}, we sum and average the linear interpolation of the $\zz^{(r)}_{t}$ and the latent conditioned on the original prompt $\PPP$,
\begin{equation}
    \label{eq:compositing}
    \zz_t \coloneqq \frac{1}{R} \sum_r w_r \times \zz_t + (1-w_r) \times \zz^{(r)}_t,
\end{equation}
$\forall t \in \{T,...,T{-}N\}$,
where the $w_r \in \Reals$ is a given weight, generally set to $0.1$ in our experiments. After $N$ steps of applying \EqRef{eq:compositing} ($N\approx 10$, see supplementary),  SD is used to refine the result and produce contextual interactions.

\subsection{Placement Finetuning}
\label{ss:pe}


In some cases the artist may wish to experiment with different object positions after obtaining a desirable image. However, when the object's bounding box is moved DD may generate a somewhat different object.
Our placement finetuning (PF) method addresses this problem, allowing the artist to immediately experiment with different positions for an object while keeping its identity, and without requiring any model fine-tuning or other optimization  \cite{Textualinversion,Dreambooth}. Note that the PF method is not intended to produce a sequence of images for a video, as that would generally require further control over 3D object pose, camera viewpoint, etc. Please refer to the supplementary for the diagram of PF.

The algorithm has three components. First, a mask $\MM_o$ for the directed object is obtained by thresholding the final $t=0$ cross-attention computed by DD
and clipping by the bounding box $\BBB$. At a selected time step $T{-}N$ the transformation $X(\cdot)$ is applied to the recorded latent and to the mask $\MM_o$, producing a translated mask $X(\MM_o)$. A mask $\neg \MM_o$ is computed as the complement of $\MM_o$ and is used to extract the background region for an initial placement-finetuned latent $\zz'_{T-N}$.


The second component (not shown in the figure) inpaints holes in the background caused by moving
the forground object. These hole regions are initialized with the transformed latent $X(\zz_T)$ masked by $\MM_o$.
In our experiments the transformation $X(\cdot)$ is translation implemented with \texttt{torch.roll}, which causes content that is translated
beyond the tensor dimension to wrap around and re-appear on the opposite side.
This has the effect of initializing the hole that would otherwise appear on the opposite side with somewhat reasonable values, however more sophisticated inpainting schemes are possible here. After these initializations
we perform a single iteration of adding noise followed by denoising, in order to cause the inpainted areas to have distinct detail and blend with the background.

Lastly, there are
$t \in \{T{-}N{-}1,\cdots,0\}$
steps in which the transformed latent from the current step is composited with the background,
\[
    \zz'_t \coloneqq \zz'_t \odot \neg X(\MM_o) + X(\zz_t) \odot X(\MM_o),
    \ifdefined\AAAISHORTEN
    \else
    \quad \forall t \in \{T-N-1,\cdots,0\}.
    \fi
\]
followed by denoising, resulting in a coherent refined image.
$N$ is generally set at $10$ in our experiments (see supplementary).
Large $N$ better preserves the original foreground and background, while smaller $N$ encourages more interaction between the foreground and background.


\section{Experiments and Comparisons}\label{sec:exp}

We now present example results of Directed Diffusion. (Please see the supplementary for additional results and comparisons).
Table~\ref{table:clip} gives the quantitative result measured by the CLIP similarities between the embeddings of the prompts and the synthesized images. 
Our results are similar to or better than the compared methods.

As described earlier, unmodified T2I methods such as \SDtext are fallible and
using such a system is \emph{not} simply a matter of typing a text prompt. A prompt such as \emph{``A dog chasing a ball''} may fail to generate the ball, generate a dog without legs, etc. This poses challenges for reporting qualitative evaluations: while there is not room to show a sufficient range of randomly chosen results, evaluating a \emph{single} randomly-chosen example often results in a failure, and is not representative of how current generative AI tools are used. Users do not simply stop when they receive a bad result, but instead do trial-and-error exploration over a number of random seeds (as well as the prompt and hyperparameters \cite{latentguide}). To emulate this, we adopt a \textbf{\SSatK} protocol, in which $k$ images are generated with consecutive seeds following an initial randomly chosen seed, and the best-of-$k$ image is subjectively selected.

\begin{table}
\centering
\begin{tabular*}{0.7\linewidth}{@{\extracolsep{\fill}} |c|c|c|c|c| }
\hline
Eval Type & SD & GLIGEN & CD & DD \\
\hline
\hline
SceneComp & 0.821 & 0.802 & 0.791 & \textbf{0.824} \\
OneMask   & 0.834 & \textbf{0.810} & - & 0.807 \\
TwoMasks  & 0.842 & 0.802 & - & \textbf{0.851} \\
\hline
\end{tabular*}
\vspace*{5mm}
\caption{CLIP scores \cite{CLIPScore,CLIP} of the three experimental categories in the \FigRef{img:comp}, \FigRef{img:mask}, and \FigRef{img:two}. See the text for details.
}
\label{table:clip}
\end{table}

\vspace*{-2mm}
\subsection{Comparison: Scene compositing}

As mentioned earlier, synthesizing and controlling several objects in an image is a challenge for many T2I methods. Common problems including missing objects, incorrectly colored objects,
and incorrectly positioned objects.


 In \FigRef{img:comp}, we showcase our results alongside the results obtained from the recommended configurations from the Composable Diffusion (CD) Colab  demo\footnote{\href{https://colab.research.google.com/github/energy-based-model/Compositional-Visual-Generation-with-Composable-Diffusion-Models-PyTorch/blob/main/notebooks/demo.ipynb}{https://colab.research.google.com/github/energy-based-model/Compositional-Visual-Generation-with-Composable-Diffusion-Models-PyTorch/blob/main/notebooks/demo.ipynb}}
 and the public implementations of BOXDIFF \cite{boxdiff} and GLIGEN \cite{GLIGEN}.
The comparison clearly reveals that the directed objects, such as the castle, cherry blossoms, church, and trees, exhibit greater visibility in our approach compared to CD.
Additionally, our results
demonstrate better fidelity to the prompt (castle with fog, pond is dark) and less information blending between objects (e.g., church and mountain are not pink).

\ifarxiv{
\begin{figure}
    \centering
    \begin{subfigure}[b]{0.48\linewidth}        
        \centering
        \includegraphics[width=1.\linewidth]{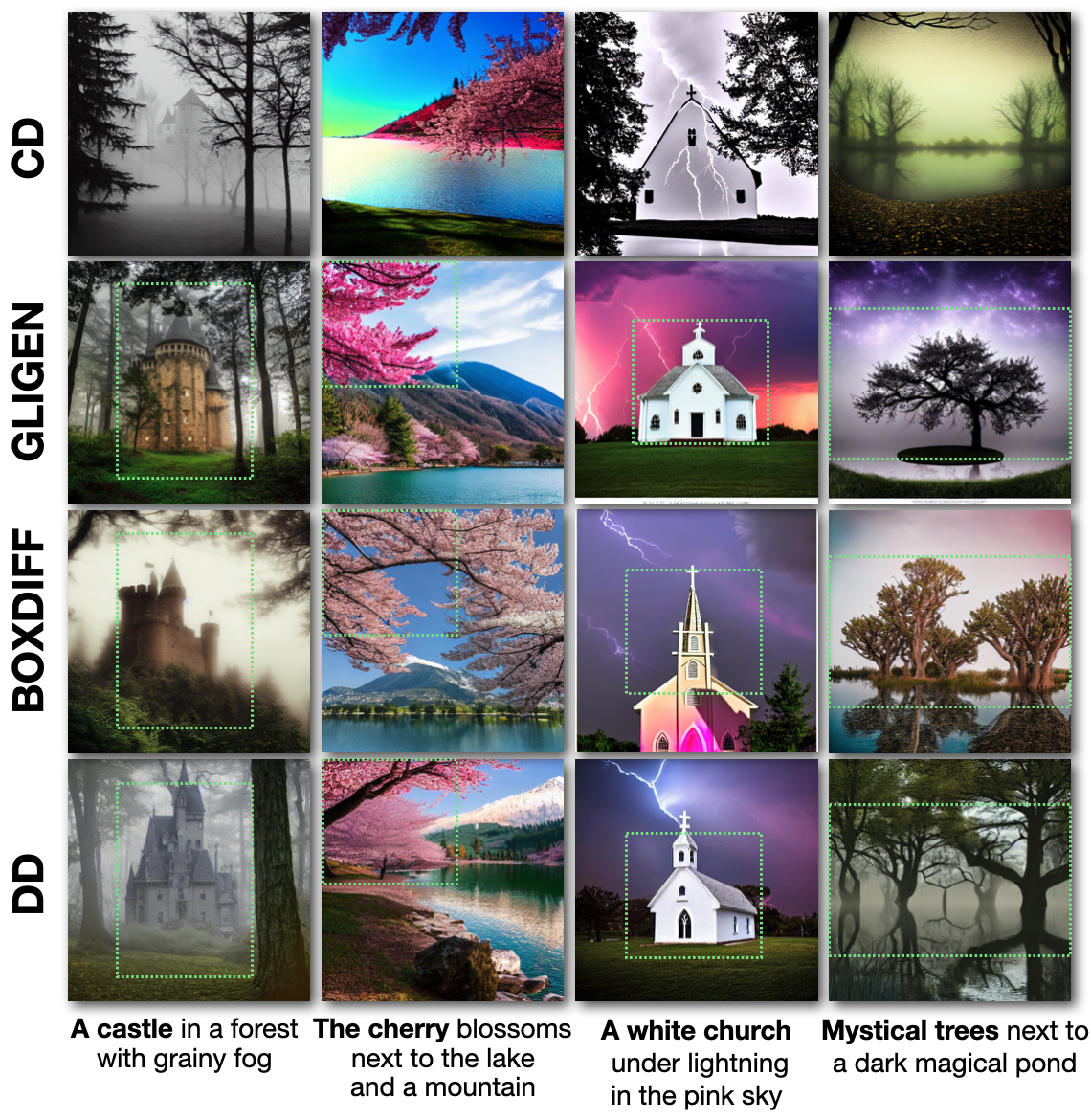}
        \caption{Composition comparison.}
        \label{img:comp}
    \end{subfigure}
    \begin{subfigure}[b]{0.48\linewidth}        
        \centering
        \includegraphics[width=1.\linewidth]{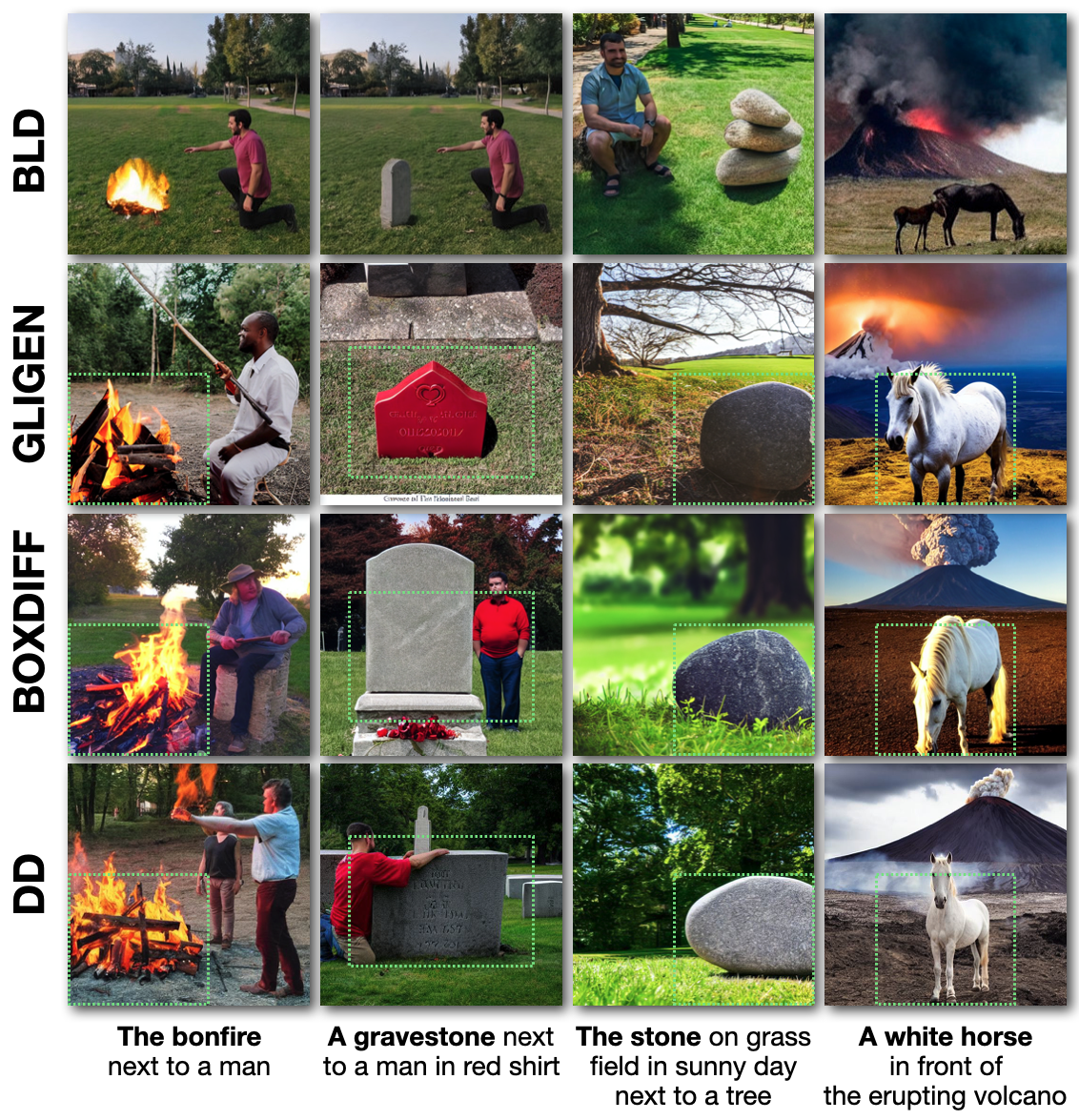}
        \caption{Masking comparison.}
        \label{fig:B}
    \end{subfigure}
    \caption{{\SSat}12 results from CD, GLIGEN, BOXDIFF, and our DD. The directed object is highlighted in bold in the prompt and its bounding box is shown in green (except for CD, which does not support position guidance). Please enlarge to see details including the bounding box. BLD images are reproduced from \cite{BlendedLatent}.}
    \label{img:mask}
\end{figure}
}
{
\begin{figure}[ht]
    \vspace{-1em}
    \captionsetup{skip=3pt}
    \centering
    \includegraphics[width=0.9\linewidth]{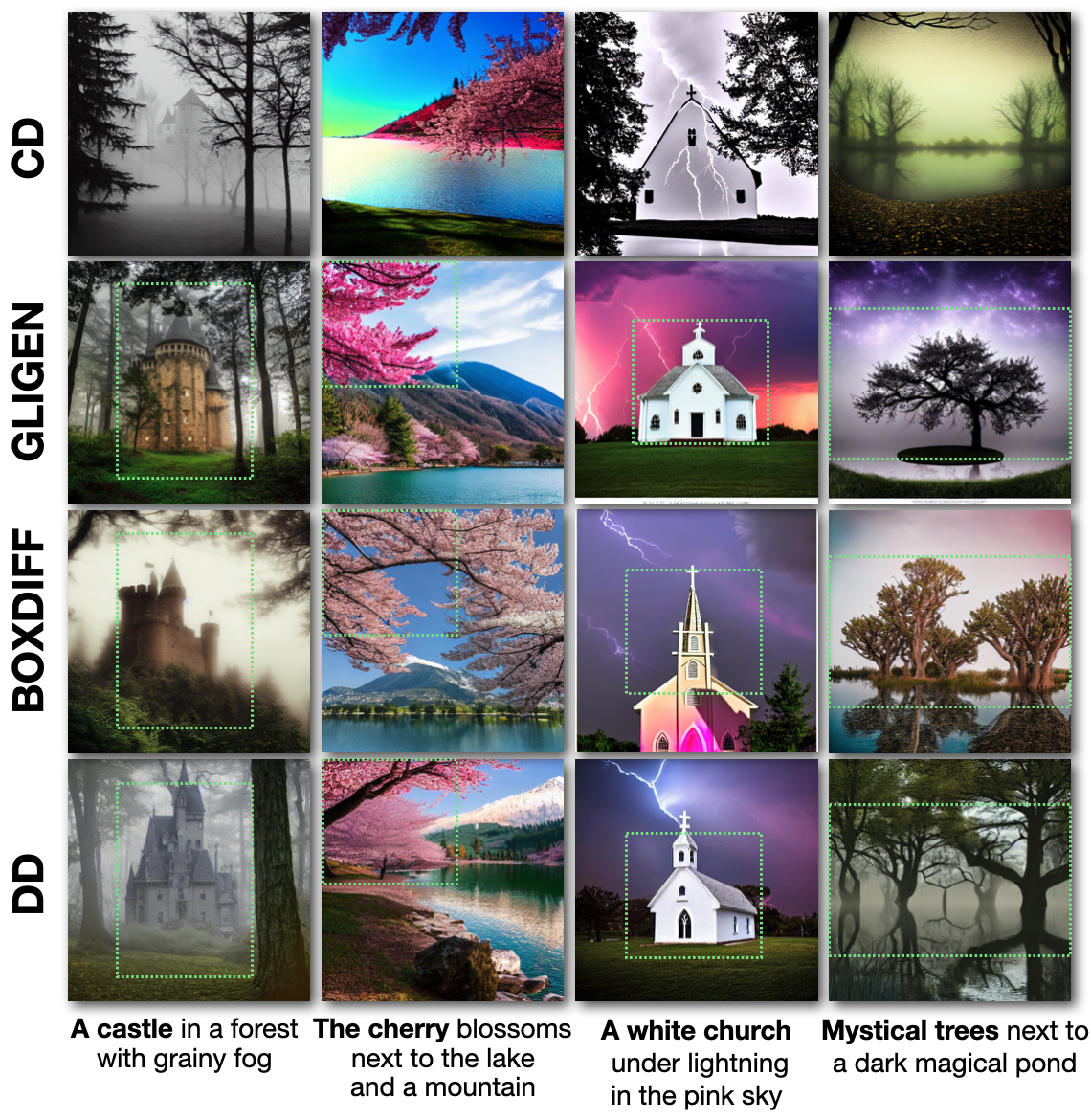}
    \caption{Composition comparison: {\SSat}12 results from
CD, GLIGEN, BOXDIFF, and our DD.
    The directed object is highlighted in bold in the prompt and its bounding box is shown in green (except for CD, which does not support position guidance). Please enlarge to see details including the bounding box.
    }
\label{img:comp}
\end{figure}

\begin{figure}[ht]
    \vspace{-0.5em}
    \captionsetup{skip=3pt}
    \centering
    \includegraphics[width=0.88\linewidth]{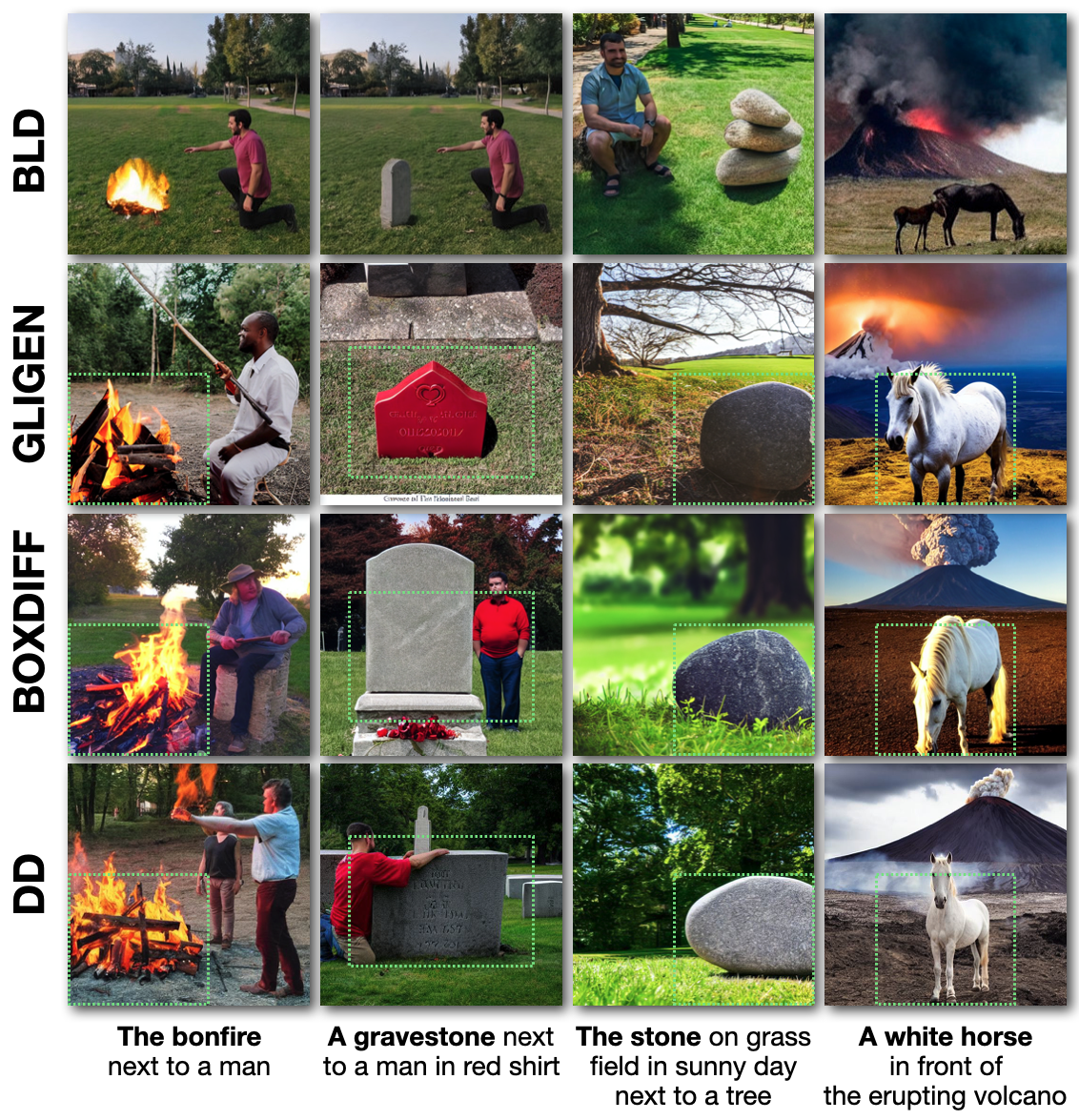}
    \caption{Masking comparison. {\SSat}12 results from BLD, GLIGEN, BOXDIFF, and our DD.
     The directed object is highlighted in bold in the prompt and its bounding box is shown in green. Please enlarge to see details including the bounding box.
     BLD images are reproduced from \cite{BlendedLatent}.}
\label{img:mask}
\end{figure}
}

\vspace*{-2mm}
\subsection{Comparison: \OneObject}

Methods for placing an object in another image can
suffer from inconsistency between the object and the background environment.
In \FigRef{img:mask}b, we compare our DD result with  Blended Latent Diffusion (BLD) \cite{BlendedLatent}, a text-driven editing method that guides object placement using using a mask. DD uses the bounding box $\BBB$ to direct the object, which can be seen as analogous to a simple form of mask.

In this comparison, note that the results shown for BLD are reproduced from the original paper, whereas we guessed a  prompt to generate roughly similar images.
In our results it is evident that DD generates strong and realistic interactions
between the directed object and the background, including
the man touching the gravestone, the shadows cast on the grass, the occlusion of the volcano by the horse, and the man's hand that catches on fire :)
We also emphasize that BLD addresses a different (and difficult) purpose of editing real images. The comparison here is intended simply to highlight the realistic interactions between the directed object and the background that are obtained using our method.

\subsection{Placement finetuning}

\FigRef{img:rolling} shows results of the placement finetuning method. \FigRef{img:rolling} (top) uses the sliding window associated with the word ``castle''. Using PF (middle), the castle identity is preserved, while the surrounding environment such as the trees and lake reflection are reasonably synthesized. The bottom row shows the reconstruction of the latent image before and after transformation. The bird is successfully re-positioned while maintaining coherent interaction with the background (E.g., reflection, waves).

\begin{figure}[ht]
    \vspace{-1em}
    \captionsetup{skip=3pt}
    \centering
    \includegraphics[width=1.\linewidth]{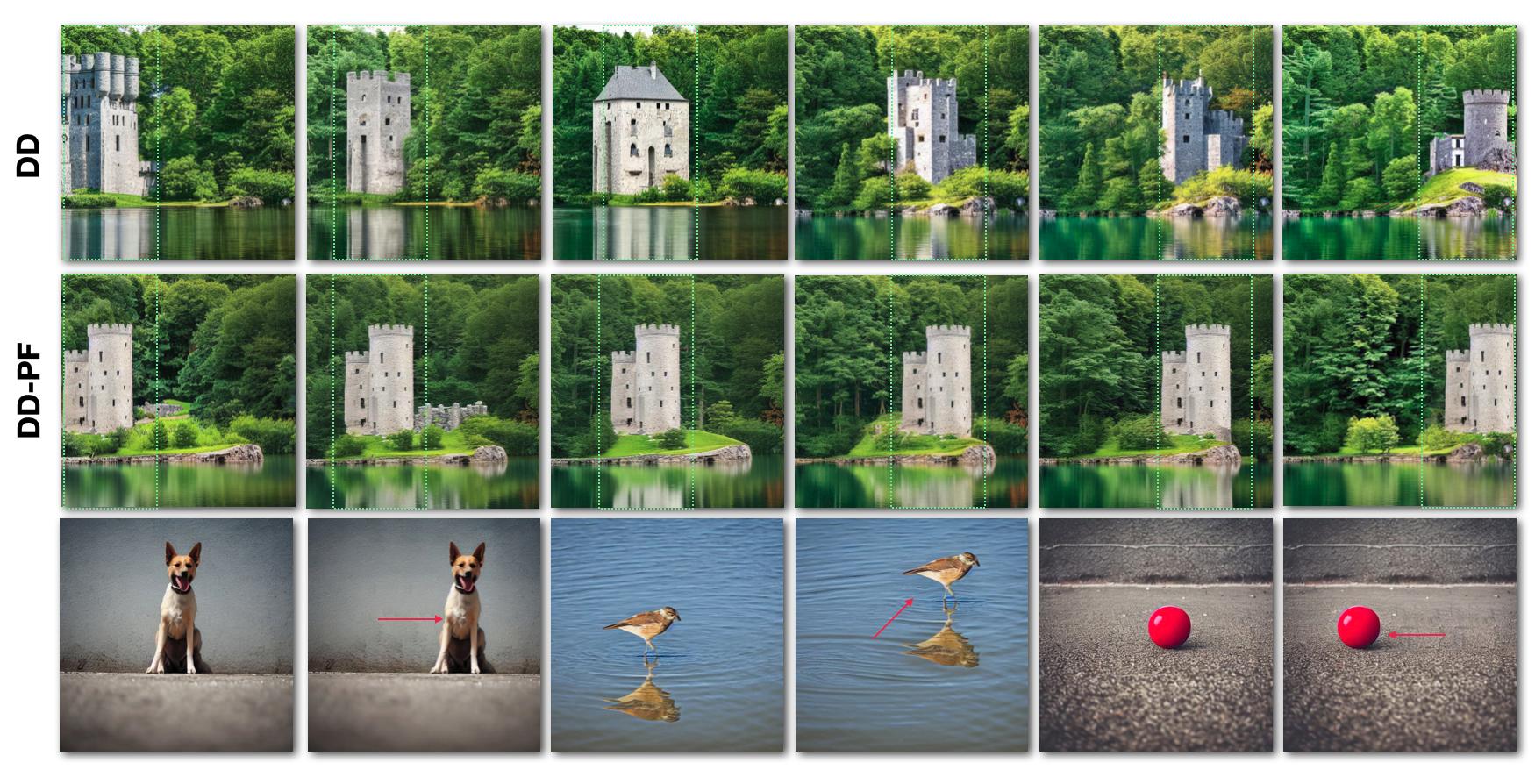}
    \caption{Comparison between DD-PF (Placement finetuning) and DD. The first row shows the DD result based on the sliding bounding box associated with castle from left to right. The second row is the PF alternative. The third row shows using PF to move the dog, bird, and sphere, respectively.}
\label{img:rolling}
\end{figure}

\subsection{Comparison: \TwoObjects}

\FigRef{img:two} compares the results of several methods in placing  two objects. In this task, a core consideration is scene consistency and the ``correlation'' between the directed objects. Our results show natural interactions such as the running dog touching the red ball, and the red cube and blue sphere supported naturally in a shopping basket.

%
\subsection{Comparison: object interactions from prompt verbs}

\begin{figure}[ht]
    \vspace{-1em}
    \captionsetup{skip=3pt}
    \centering
    \includegraphics[width=1.\linewidth]{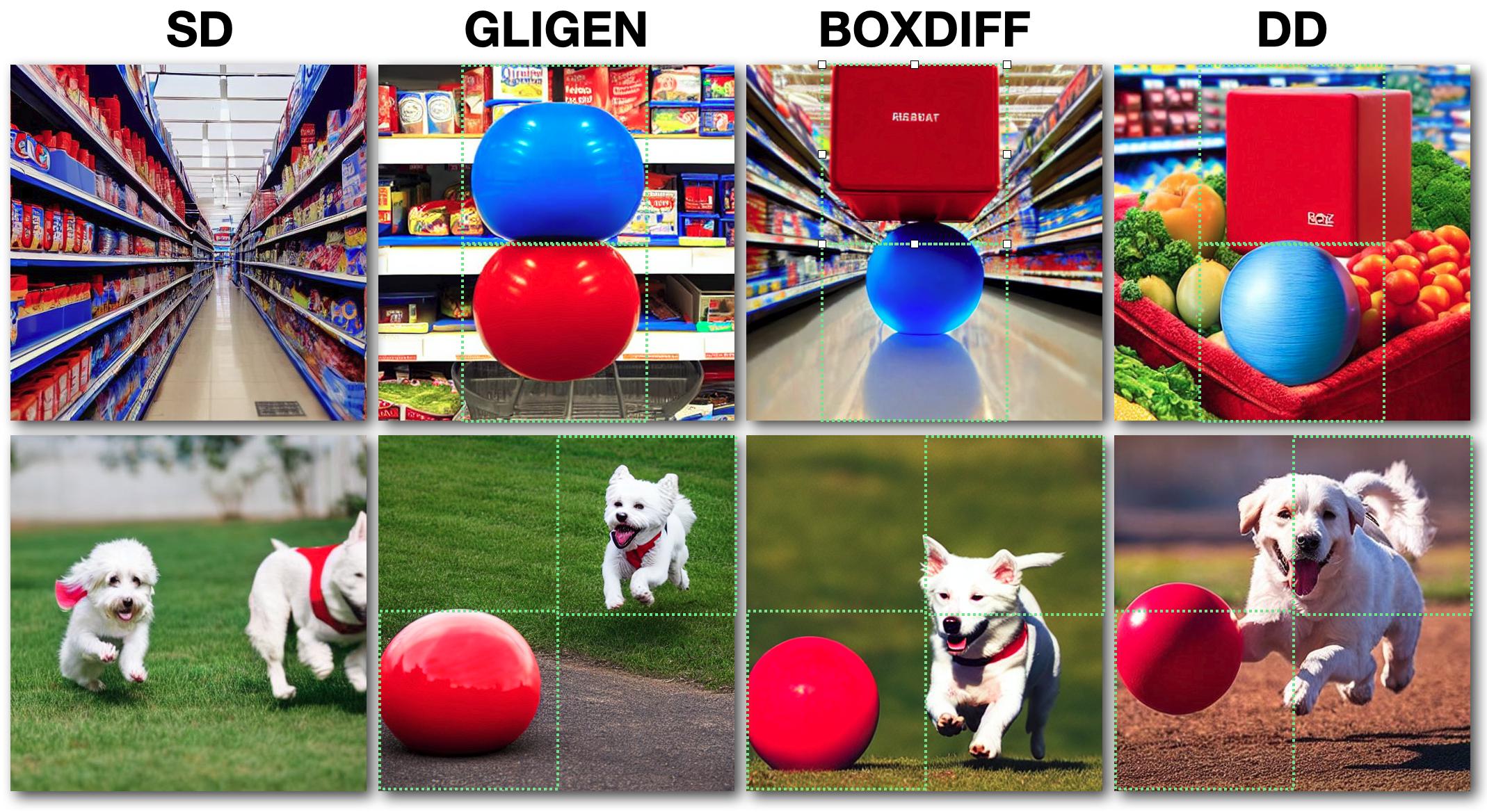}
    \caption{{\SSat}12 results from SD, GLIGEN, BOXDIFF, and DD. The prompts are (first row) ``A \textbf{red cube} above \textbf{blue sphere} in the supermarket'', (second row) ``A \textbf{white running dog} chasing after a \textbf{red ball}.''
    The directed object is denoted in bold and its bounding box is shown in green. Please enlarge to see details including the bounding box.
    }
\label{img:two}
\end{figure}

\begin{figure}[ht]
    \vspace{-1em}
    \captionsetup{skip=3pt}
    \centering
\includegraphics[width=1.\linewidth]{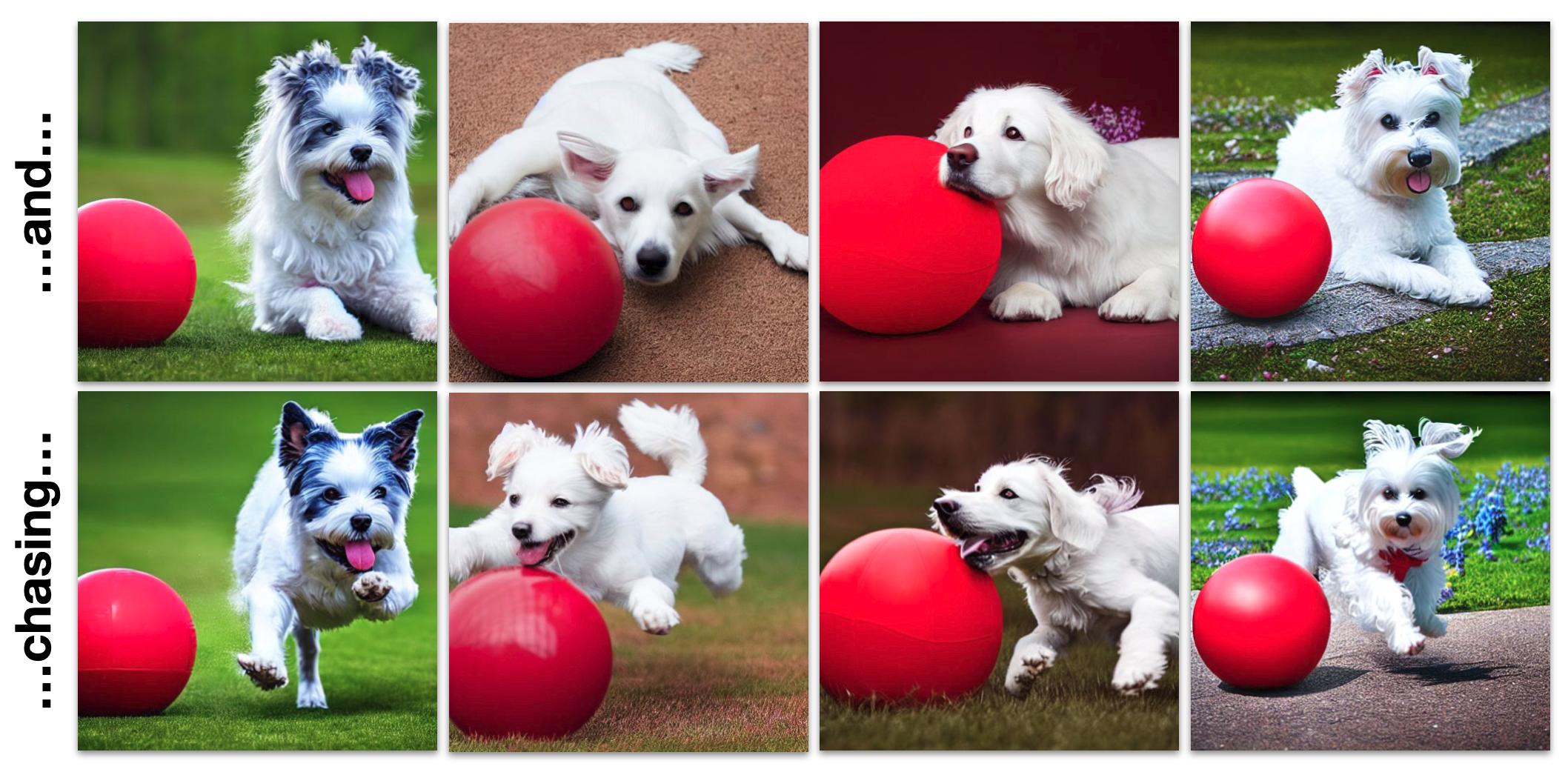}
    \caption{{\SSat}12 results of the prompt \emph{``a white dog S* a red ball''}, where the token S* is replaced by \emph{``and''} or  \emph{``chasing''} in the first row and second row, respectively. In the second row the dog more frequently appears to be running rather than sitting, consistent with the verb ``chasing''. Note the results of each column share the same random seed.}
\label{img:chasing}
\end{figure}

\FigRef{img:chasing} showcases the ability of DD to (sometimes) convey interaction between objects. The figure compares two prompts: \emph{"A white dog \textbf{and} a red ball"} and \emph{"A white dog \textbf{chasing} a red ball."}
While both results show the white dog and red ball consistently,
it is evident that the ``chasing'' prompt better conveys the action of this verb, e.g.~the dog is running rather than sitting. Note however that this capability is bounded by CLIP's limited understanding of language structure \cite{TrainingFreeStructuredDiffusion} and will certainly fail with more complex sentences.


\section{Limitations and Conclusion}

\ifdefined\AAAIOURVERSION

Storytelling with images requires directing the \emph{placement} of important objects in each image.
Our algorithm, directed diffusion, is a step toward this goal.
The algorithm is simple to implement, requiring only a few lines of  modification of a widely used library \cite{diffusers}.
Directed Diffusion inherits imitations of \SDtext, including the need for trial-and-error exploration mentioned earlier.
Some random seeds fail to produce the desired subject or produce distorted objects such as animals with missing limbs.
In common with methods such as \cite{SDEdit,Magicmix}, it is necessary to specify the number of steps over which editing is active.
Objects do not strictly respect the provide bounding box and may extend outside it. On the other hand, this is an advantage as well as a disadvantage, since forcing the object to closely follow the box can result in an unnatural appearance (Fig.~\ref{img:trailing} in supplementary).
Significant additional advances will be needed before video storytelling is possible.
However, our method may be sufficient for the creation of storybooks, comic books, etc.~when 
used in conjunction with other existing tools.

\else
This paper continues a graphics research tradition in which an artist is guiding and interacting with generative tools.
On the other hand, recent advances in generative models (of both images and text) suggest that a future in which anyone can make a movie using only high level direction ("make me a comedy movie about a cat that saves the world by ...") is no longer science fiction. Several factors suggest
this future is still distant, however.  For one, the overall generation of stories is not a fully solved problem. This includes aspects such as the narrative arc and synthesis of the film language needed to convey the story to a viewer.
Second, neural video generation experiments generally trade spatial for temporal resolution due to memory limits, and current results generally show a few seconds of video at fairly low resolution (recall that mainstream movies are 4K resolution), while sometimes suffering from flickering and distorted objects.
In conclusion, there remains a role for artist-directed graphics in the near future.
\fi

\section{Acknowledgements}
We thank Jason Baldridge for helpful feedback.

\nocite{paszke2017automatic, kdiffusion, Karras2022edm}
\bibliographystyle{unsrtnat}
\bibliography{main}

\clearpage

This supplementary presents general \textbf{implementation details}, further details on the implementation of placement finetuning and scene compositing, an \textbf{ablation} on the core optimization method, more extensive \textbf{qualitative comparisons} against several alternative methods, several further \textbf{quantitative evaluations} along with a discussion of their \textbf{limitations}. Following the evaluations we give several comments on how the reproducibility guidelines are addressed in the paper.

\section{Implementation} \label{sec:imp}

Here we discuss our implementation in more detail. DD is implemented with PyTorch 2.01 \cite{paszke2017automatic} and the \textsc{diffusers} library version 0.14 from Huggingface \cite{diffusers}. We use the pretrained stable diffusion model CompViz/stable-diffusion-v1-4 \cite{RombachStablediffusion21} for all our experiments. The classification free guidance scale is 7.5. The Linear Multistep Scheduler implemented in diffusers are used for discrete beta schedules based on k-diffusion \cite{kdiffusion, Karras2022edm}. The examples used 50 denoising steps.
The default PyTorch random generator was used. The seed was initialized at 0 (1 in the case of BoxDiff and incremented and incremented through each experiment. The experiments were performed on a single Linux OS machine containing a consumer NVIDIA 3090 GPU (24GB memory). Generating a 512x512 image takes roughly 10 seconds on this machine.

The attention editing stage of the DD pipeline (Fig.~3 and \Eq~1 in the main text), and the scene compositing (Scene compositing section and \Eq~2 in the main text),
both involve a choice of the number of editing steps $N$.
This choice is informed by the rate of change of the predicted noise between the steps. \FigRef{img:normgrad} illustrates the norm $\|\nabla_t \zz_t\|$ of the gradient of the latent with respect to the time steps.

As can be seen, there is an initial period of $\approx 10$ time steps labeled as 40 in x-axis in which the norm is rapidly decreasing, indicated by a yellow arrow in the figure. During this period the initial shape of objects are established. This is seen in \Fig 2 in the main text, which visualizes reconstructions from the VAE decoder $\DDD(\cdot)$ and the cross-attention maps $\bAA^{(i)}, \forall i \in \{1,...,|P|\}$ at several time steps during the denoising.

Following this initial period, there is a subsequent stage in which the latents change more slowly as fine details are filled in (visualized with the green arrow). We run the editing operations during the initial ``yellow'' phase, while conventional SD refinement is used in the ``green'' phase.
For nearly all experiments we empirically selected $N{=}10$ (indicated with the star in the figure) as the time step to switch from editing to refinement.

\begin{figure}[H]
    \centering
    \includegraphics[width=0.5\linewidth]{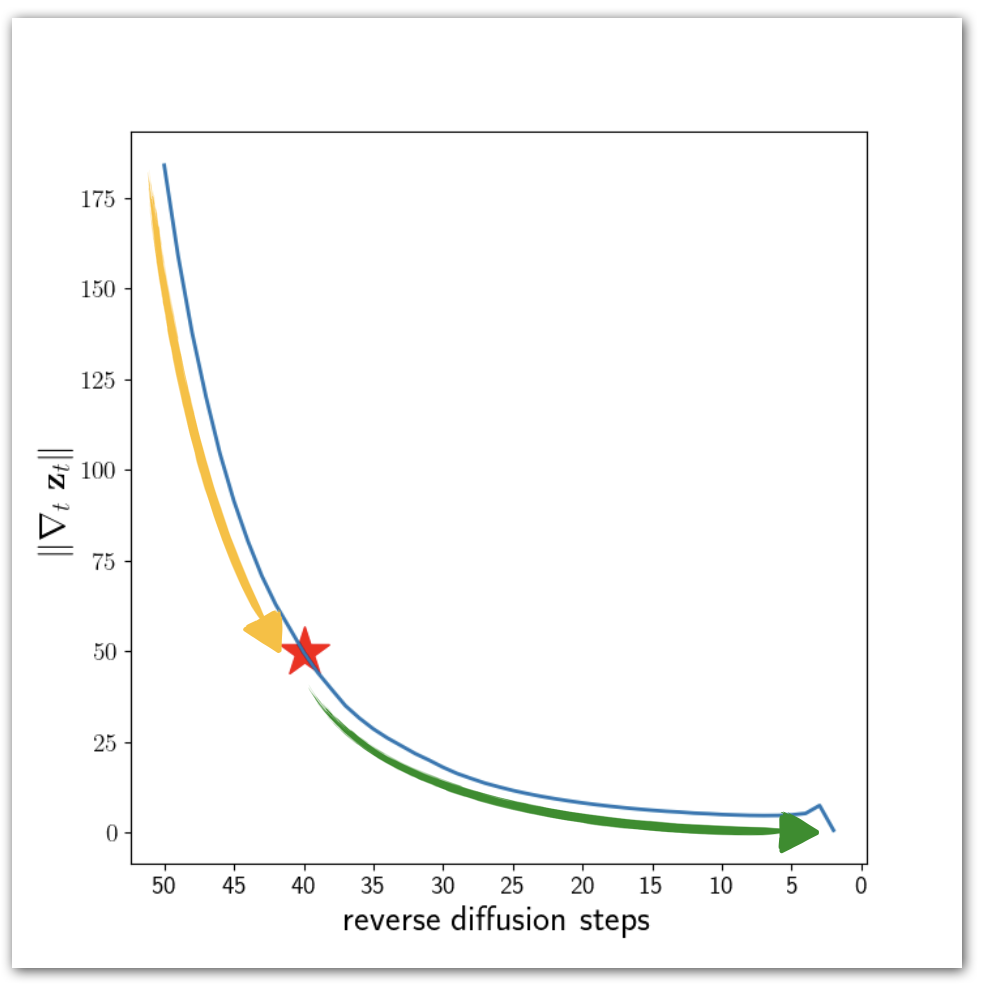}
    \caption{The norm of the gradient of latents with respect to time, plotted across time steps. The yellow and the green arrows indicate two different phases that correspond to our editing and refinement stages.}
    \label{img:normgrad}
\end{figure}

The optimizer Adam is used with the learning rate 0.0005 to optimize the Eq.(1) in the first stage. The vector $\aa$ is initialized with a uniform distribution in the interval $[0,c]$, where generally $c{=}0.15$ in our reported experiments. A higher constant tends to lead to an empty background, while very low values fail to guide the directed object.


\section{Analysis and Ablation} \label{sec:trailing}

Several concurrent papers direct object placement by manipulating the cross attention from tokens corresponding to the directed objects. DD adopts an unusual approach of directing objects by doing a small optimization to reweight ``trailing attention maps''. Here we give further intuition on this choice and present evidence in the form of an ablation where the optimization is removed.

A number of text-to-image models use CLIP \cite{CLIP} embeddings for the text guidance. CLIP accepts an input text representation with up to 77 tokens, including the initial ``[CLS]'' token representing the sentence level classification. The number of tokens $|\mathcal{P}|$ in the prompt $\PPP$ is generally less than this number. We call the remaining $\TTT \in \{k | |\PPP| < k \leq 77 {-} |\PPP| {-} 1\}$ attention maps corresponding to the non-prompt tokens the \emph{trailing attention maps} $\bAA^{(i)}, \forall i \in \TTT$. Note that the output embedding for [CLS] is not used in our approach.

\begin{figure}[H]
    \centering
    \includegraphics[width=0.8\linewidth]{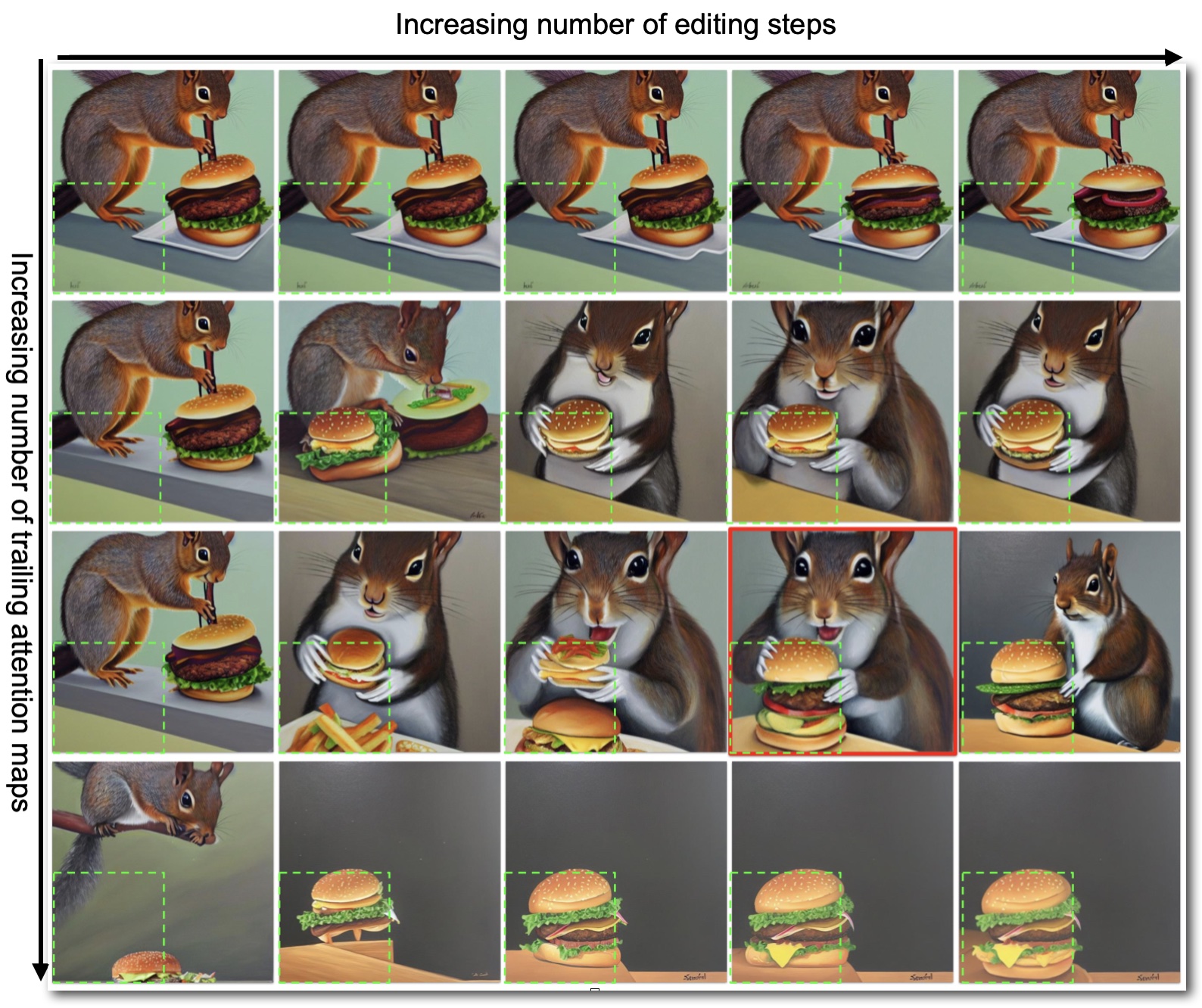}
    \caption{The number of trailing attention maps (5, 10, 15, 20, maps on the vertical axis) versus the number of attention map editing steps (1, 3, 5, 10, and 15 steps on the horizontal axis). The prompt is \emph{``A photo of a squirrel eating a burger''} with the directed object ``burger'' positioned at the bottom left. The best results are obtained with an intermediate number of editing steps and edited trailing attention maps, as in the case of the image with the red border.
    }
    \label{img:trailing}
\end{figure}

We edit the trailing maps by injecting Gaussian-falloff activations inside the user-specified bounding box, followed by a small optimization (Eq.~1 in the main text) to find best weights for the trailing maps, resulting in the targets $\DD^{(i)}$ in section~3.2 and Algo.~1 in the main text.

{
It may not be initially obvious why the trailing attention maps should affect object placement, nor why we would want to use them.
While the \textsc{diffusers} library implements multi-head attention and a variety of conditioning mechanisms, the effect of the trailing maps can be seen from considering a generic cross attention scheme.
The cross attention maps $\bAA^{(i)} = \text{Softmax}(\QQ {\KK^{(i)}}^T), \forall i \in \{1,\cdots,\NW \}$
where $\QQ \in \Reals^{n^2 \times L}$
and $\KK \in \Reals^{\NW \times L}$,
are essentially the cross-covariance of each ``pixel'' of the projected embedding of the \SD latent $z_t$ with projected embeddings of the $\NW=77$ potential words from the prompt.
Consider the standard cross-attention computation $\CC = \text{Softmax}(\QQ \KK^T) \cdot \VV$,
where $\VV \in \Reals^{\NW \times L}$ is a second projected text embedding of dimension $L$ for each token. The resulting conditioning matrix $\CC$ has dimension $\Reals^{n^2 \times L}$, while a single ``pixel'' of this matrix is a convex sum (from the softmax) of all the embeddings in $\VV$, \emph{including those in the trailing words.}
} 

Manipulating the trailing maps thus influences the conditioning used in the (classifier-free) text guidance. The optimization scheme in \Eq~1 in the main text uses trailing maps to provide a clean separation between the desired object location (specified via the trailing maps at time step $t$) and the computed attention map at time $t{-}1$.
As stated in the main text, we believe that this intermediate denoising step serves the purpose of moving our position guidance near the range of values expected by the pretrained SD model.
The optimization exploits the fact that the  maps do not have identical effects (ultimately due to the random initialization of the model), to find a weighted combination that successfully produces the desired object direction.

\FigRef{img:trailing} shows an ablation in which the optimization in Eq.~1 is removed and the Gaussian-falloff activations are instead directly injected in the cross attention maps for the directed prompt tokens as well as a specified number of trailing maps. The figure shows the number of trailing maps plotted versus the number of attention editing time steps. When the number of edited trailing attention maps is small (e.g., first row, $|\mathcal{T}|=5$), the image $\xx_0$ is barely modified. In contrast, when the number of edited trailing attention maps is too large (last row), $|\mathcal{T}|=20$), $\xx_0$  is missing the background and satisfies only the bounding box information. Using an intermediate number of edited trailing attention maps results in the image simultaneously portraying the desired semantic content at the desired location(s).

We initially experimented with implementing DD by directly injecting the Gaussian-falloff activations in the cross-attention maps for the directed prompt words and a specified number of trailing attention maps, however (as can be seen with this ablation) this   was somewhat unstable and required a grid search to obtain good results. To avoid this, we instead formulate \Eq 1 to automatically obtain the best combination of the trailing attention maps to control the directed object.

Here we show more experiments with one directed object with a guided bounding box positioned in the four quadrant and the slide window style.

\FigRef{app:img:four-q} shows examples of directing objects to lie in each of four image quadrants as well as at the image center, with the directed objects ``a large cabin'', ``an insect robot'', and ``the sun''. Notice for example that the ``sun'' examples show the effect of the sun in the correct quadrant but in different contexts. The 3rd quadrant and the center show the real sun at the edge of the house roof, while the other images show the reflection of the sun such as from a window (e.g., 1st, 4th quadrant), or the wall (e.g., 2nd quadrant).

\FigRef{app:img:sliding} shows the result of moving an object from left to right. All the results are generated with a bounding box with the full image height and 40\% of the image width. The directed objects are ``a stone castle'', ``a white cat'', and ``a white dog'', respectively.

\begin{figure}[H]
    \centering
    \includegraphics[width=1.0\linewidth]{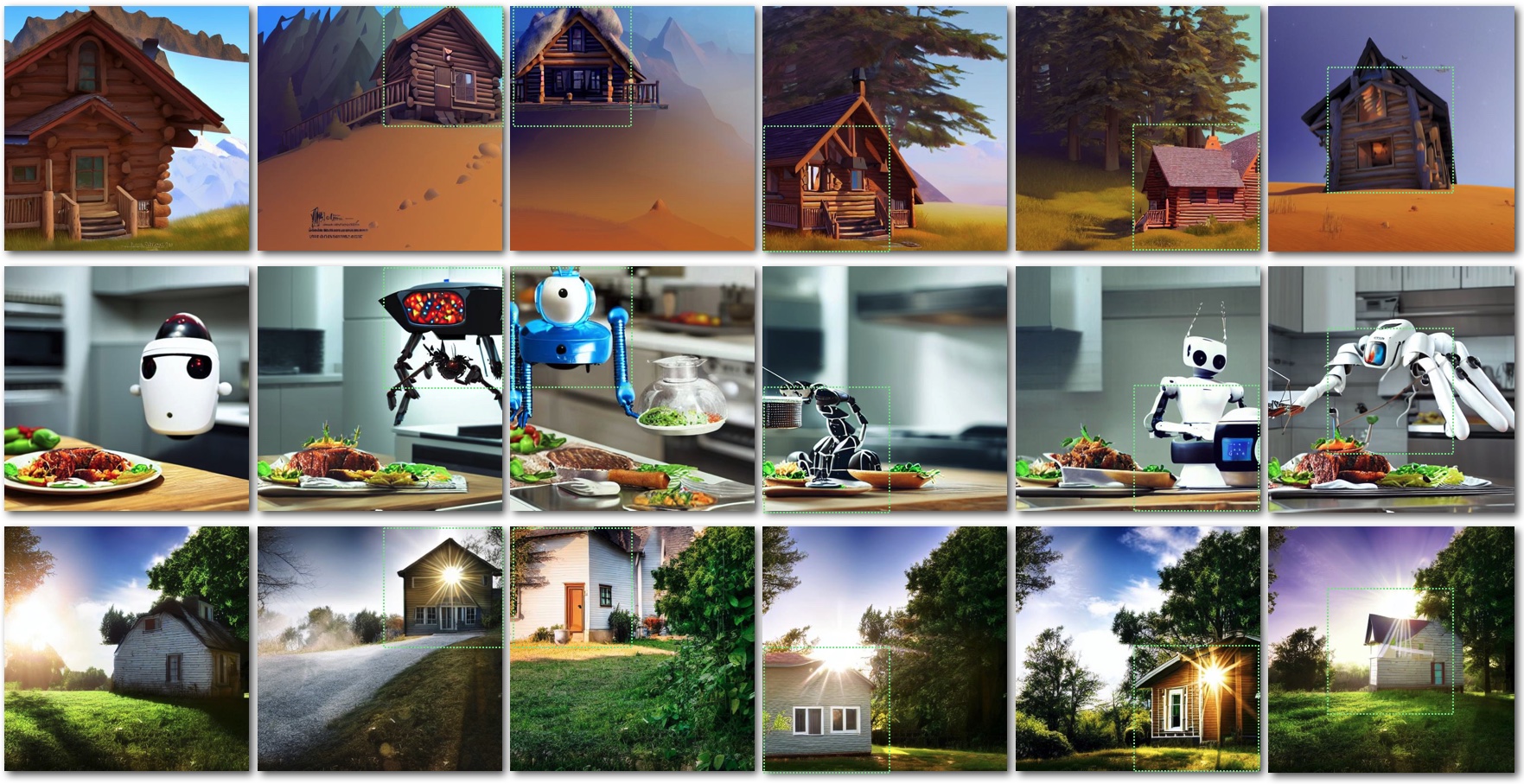}
    \caption{Four quadrants: (Top) Prompt: \emph{``A large cabin on top of a sunny mountain in the style of Dreamworks artstation''}, with the directed object ``a large cabin''. (Middle) Prompt: \emph{``An insect robot preparing a delicious meal and food in the kitchen''}, with the directed object ``an insect robot''. (Bottom) Prompt: \emph{``The sun shines on a house''}, with the directed object ``the sun''. Images from the left are \SD, DD(1st Q), DD(2nd Q), DD(3rd Q), and DD(4th Q), respectively, where Q denotes the directed quadrant.}
    \label{app:img:four-q}
\end{figure}

\begin{figure}[H]
    \centering
    \includegraphics[width=1.0\linewidth]{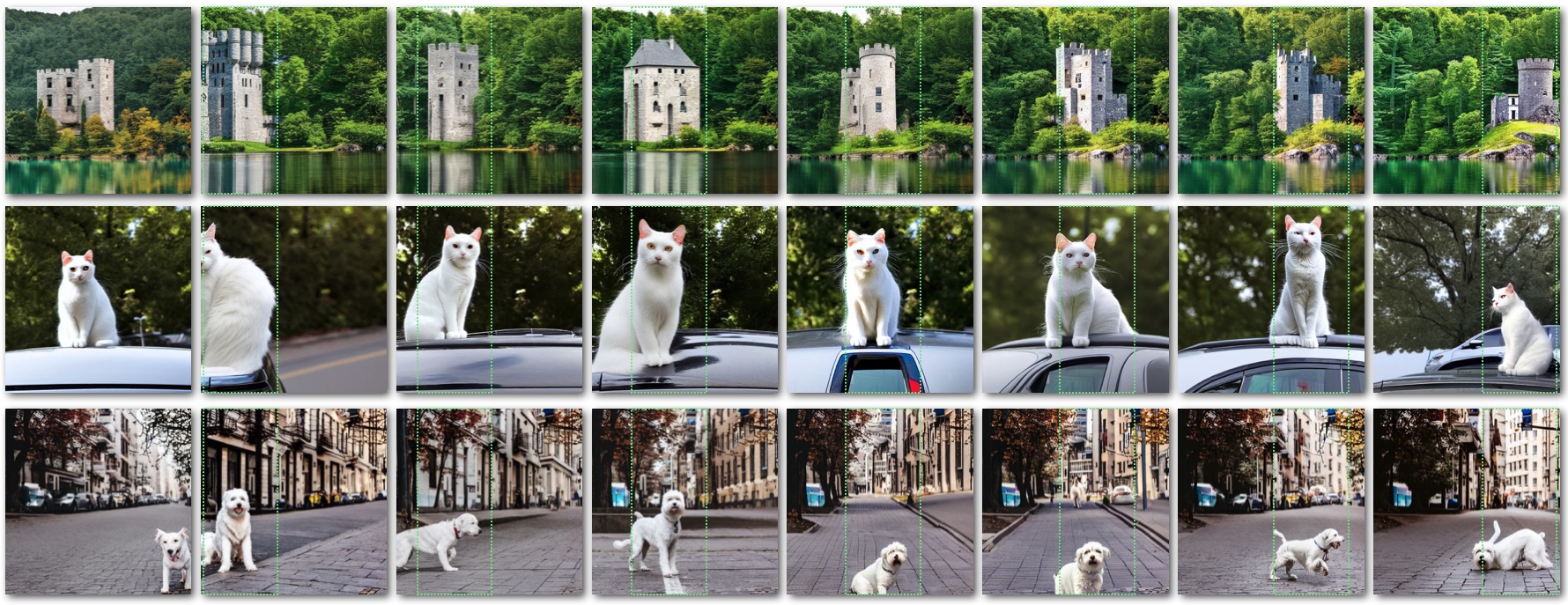}
    \caption{Sliding window: (Top) Prompt: \emph{``A stone castle surrounded by lakes and trees''}, with the directed object ``A stone castle''.
    (Middle) Prompt: \emph{``A white cat sitting on a car''}, with the directed object ``A white cat''.
    (Bottom) Prompt: \emph{``A white dog waking on the street in the city''}, with the directed object ``A white dog''. From the left: SD, $\text{DD}_h(0\%,40\%)$, $\text{DD}_h(10\%,50\%)$, $\text{DD}_h(20\%,60\%)$, $\text{DD}_h(30\%,70\%)$, $\text{DD}_h(40\%,80\%)$, $\text{DD}_h(50\%,90\%)$, $\text{DD}_h(60\%,100\%)$, where $\text{DD}_h(\cdot,\cdot)$ denotes the denotes the left and right edge of the directed region, expressed as percentages of the image resolution.}
    \label{app:img:sliding}
\end{figure}

\section{Placement finetuning}

Placement finetuning offers a way to refine the DD (or, conventional stable diffusion, or any diffusion-based method) result without changing the subject content while keeping the environment context consistency. \FigRef{img:post} below shows a diagram of the whole process to supplement our main paper.

\begin{figure}[th]
    \centering\captionsetup{skip=3pt}
    \includegraphics[width=0.8\linewidth,trim=0 0 0 0,clip]{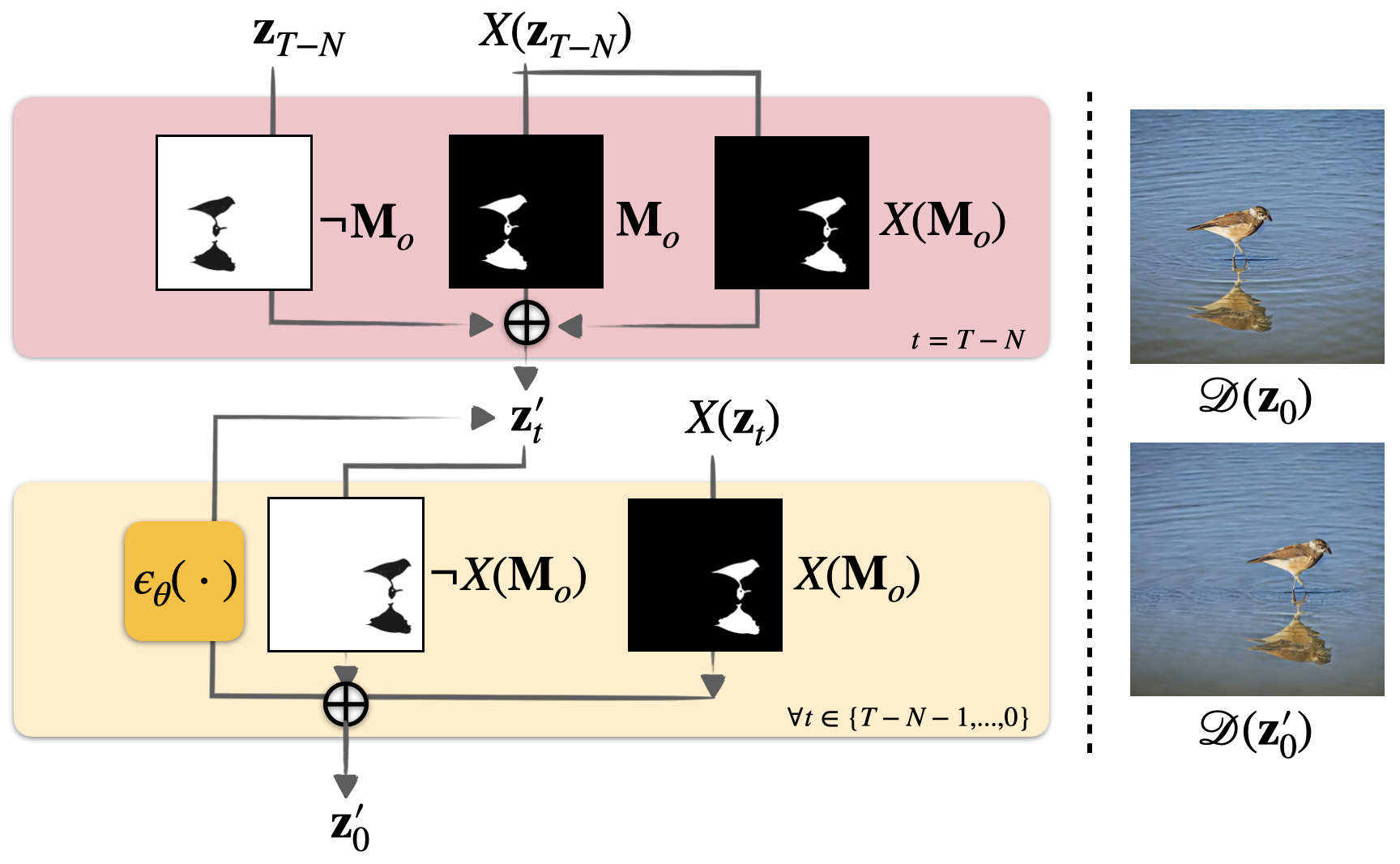}
    \caption{Placement finetuning.
     The final (t=0) cross-attention map recorded from a previous DD synthesis is segmented into foreground and background. At a chosen editing time step $T-N$ the foreground is moved and the background is inpainted. This edited latent is then further denoised to produce a coherent image. See the text for details.
    }
    \label{img:post}
\end{figure}

We first define three masks by using the cross attention map of the directed object in the prompt (e.g., bird), namely, the background mask $\neg \mathbf{M}_o$, the foreground mask before moving the object $\mathbf{M}_o$, and the foreground mask after moving $X(\mathbf{M}_o)$, where $X(\cdot)$ is the spatial transformation function (restricted to translation in our current implementation). $\neg \mathbf{M}_o$ is used to inpaint the area after the object is moved, and $\mathbf{M}_o$, $X(\mathbf{M}_o)$ extract the shifted latent $X(\mathbf{z}_{T-N})$ to get the latent of the transformed object. This process is only executed once at denoising time step $T{-}N$.
Finally, it modifies the latent to keep the repositioned directed object as constant as possible 
while producing a seamless blend with the background by denoising the composited result into $\mathbf{z}'_t$.

\section{Scene compositing}

\FigRef{img:compositing} provides an additional diagram of scene compositing to supplement the description in the main paper.

\begin{figure}[H]
  \vspace{-1em}
    \centering\captionsetup{skip=3pt}
    \includegraphics[width=0.8\linewidth]{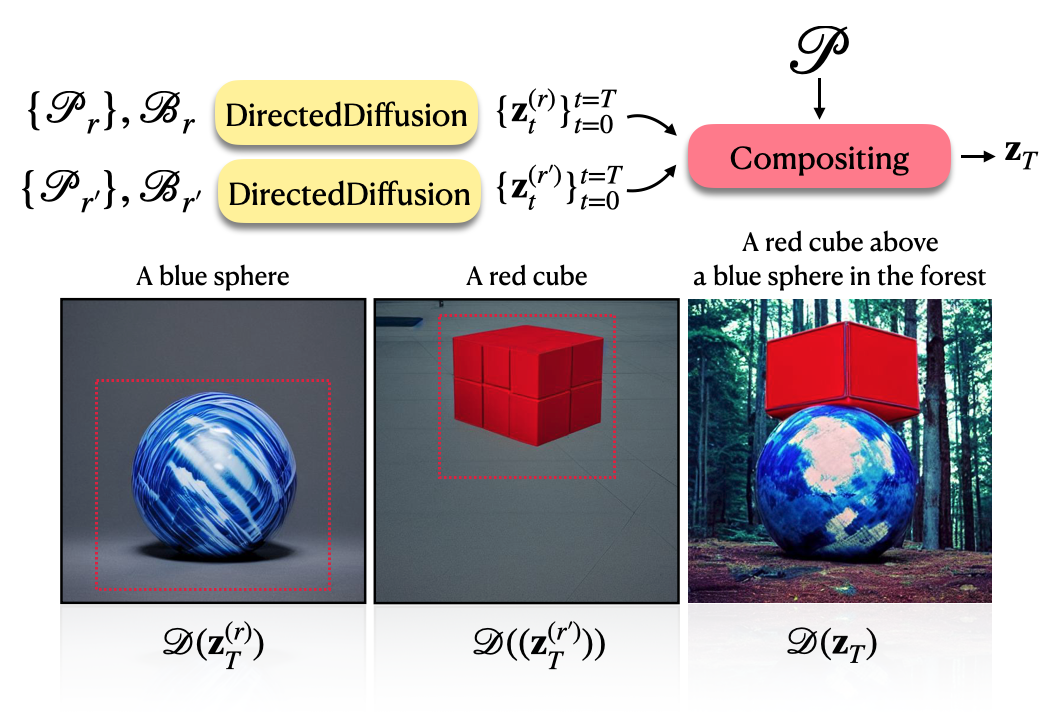}
    \caption{Scene compositing. The synthesized reconstruction $\mathcal{D}(\zz_T)$ takes multiple latent results from DD (e.g., $\zz^{(r)}_T$ and $\zz^{(r')}_T$) and the original prompt $\PPP$, whose tokens are a superset of those of the individual sub-prompts ($\PPP_r$, $\PPP_{r'}$ in this example).}
    \label{img:compositing}
\end{figure}
To sum up, DD first generates directed objects separately (e.g., a blue sphere, a red cube). For illustration purposes, the first two figures on the left depict the reconstructed sphere and cube. Subsequently, we engage in a compositing step to combine the latents of each directed object $\mathbf{z}^{(r)}$, taking their respective bounding boxes into account.
After completing $N{=}10$ compositing steps (see main paper) the resulting $\zz_t$ is then denoised with conventional SD.

\FigRef{app:img:2boxes} shows two experiments each with two directed objects: \FigRef{app:img:2boxes} (top) directs the objects \emph{``red cube'', ``blue sphere''}, while \FigRef{app:img:2boxes} (bottom) shows \emph{``bear'', ``bird''}, respectively. This is the well known ``compositionality'' failure case for SD,\footnote{This failure case is highlighted on the huggingface \cite{huggingface} stable diffusion website  \url{https://huggingface.co/CompVis/stable-diffusion-v-1-1-original}}, however, DD is able to handle this case. Additionally, unlike \cite{LiuCompositional22}, which addresses this problem using a specifically trained model in their polygons example, we are able to use the generic pre-trained \SD model without further modification. This results in our backgrounds having a variety of natural textures correctly aligned with the prompt contextual information (for example, the shadow of the blue sphere). In addition, the correlation of the directed objects is also retained, such as the \emph{``watching''} prompt word associated with the bear and the bird.

\begin{figure}[H]
    \centering
    \includegraphics[width=1.0\linewidth]{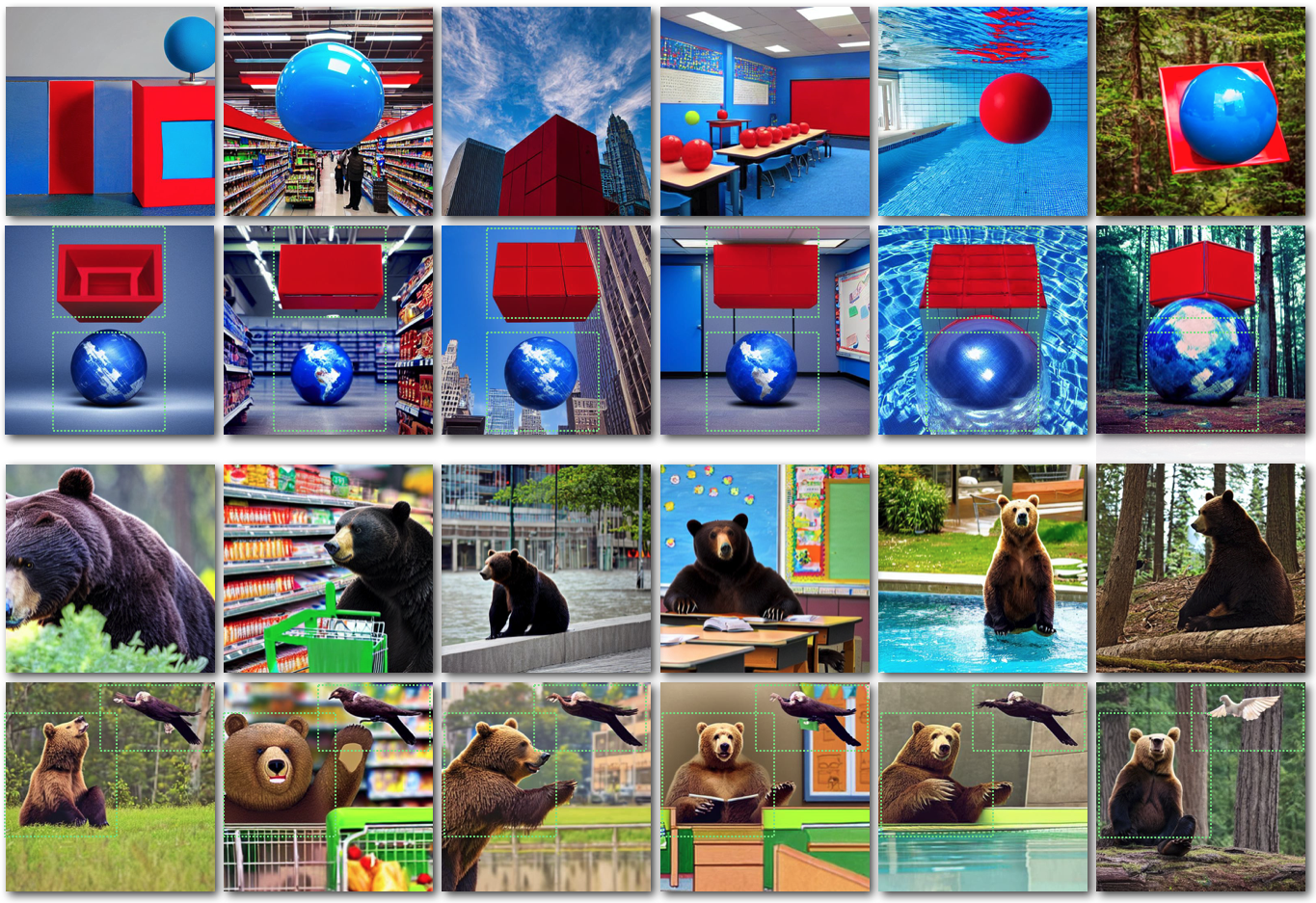}
    \caption{Two directed objects. The second and fourth rows show DD directing two objects with the green bounding boxes. For comparison, the first and third rows show the SD results on the same prompt. The first column is the original prompt \emph{``A red cube above the blue sphere''} and \emph{``A bear watching a bird''}, respectively, followed by columns appending various contexts at the end of the prompt: \emph{``in the supermarket''}, \emph{``in the city''}, \emph{``in the classroom''}, \emph{``in the swimming pool''}, and \emph{``in the forest''}.}
    \label{app:img:2boxes}
\end{figure}

\begin{figure}[H]
  \centering
    \centering\captionsetup{skip=3pt}
    \includegraphics[width=1.\linewidth]{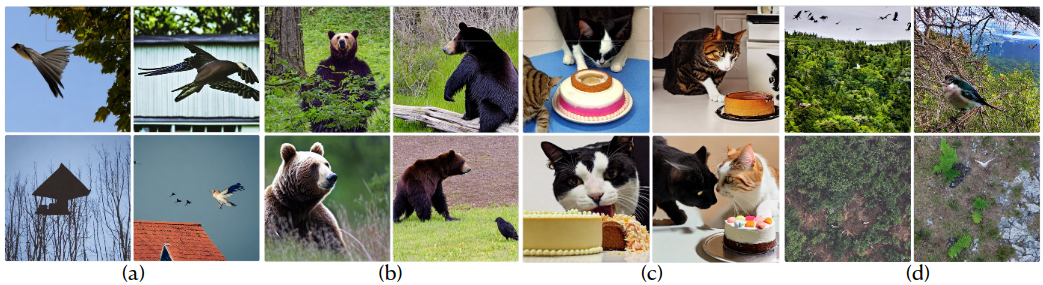}
    \caption{Stable Diffusion \cite{huggingface} failure cases.
    (a) \emph{A bird flying over a house}: no house, flying house. 
    (b) \emph{A bear watching a bird}: no bird, bear is not watching bird.
    (c) \emph{A cat watching a dog eat cake}: cat is eating, no dog.
    (d) \emph{A nature photo with a bird in the upper left}: ``upper left'' is ignored.}
   \label{img:sd-prob}
\end{figure}


\begin{figure}
    \centering
    \ifarxiv{\def\thiswidth{0.5\linewidth}}{\def\thiswidth{0.8\linewidth}}

    \includegraphics[width=\thiswidth]{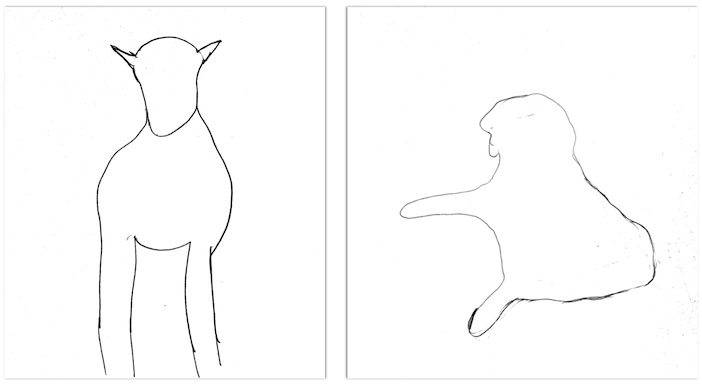}

    \vspace{-0.5em}
    \caption{Users who are not artists have difficulty producing realistic detailed masks for placement guidance. Left: attempted mask for \emph{``A camel walking toward you''}. Right: \emph{``A dog jumping with a ball''}.
    }
    \label{fig:usersketch}
    \vspace{0.5em}
\end{figure}

\section{Qualitative Evaluations}

This section provides a more extensive comparison of our method DD with the counterparts BoxDiff \cite{boxdiff} and GLIGEN \cite{GLIGEN}, together with a brief comparison to Multidiffusion \cite{multidiffusion}.
\footnote{We reproduce the result with the authors' public repositories GLIGEN:\url{https://github.com/gligen/GLIGEN} @commit:f0ede1e; BoxDiff:  \url{https://github.com/showlab/BoxDiff} @commit:3dbedc5; Multidiffusion: \url{https://huggingface.co/spaces/weizmannscience/MultiDiffusion}
}
Although these methods were concurrently developed and we only recently became aware of them, to our knowledge they are the methods that satisfy the criteria laid out in the related work section, i.e., excluding methods that require training for layout guidance (e.g.~\cite{RombachStablediffusion21}), fine tuning, or the use of detailed masks (\FigRef{fig:usersketch}).

An issue in showing qualitative comparisons is the fact that T2I methods often simply fail to synthesize the objects requested by the prompt, with successful results only being obtained after experimentation.
\FigRef{img:sd-prob} provides some examples of failures in the case of prompts that involve multiple objects. Even with single objects, however, \SD often fails, for example, objects may have incorrect colors \cite{TrainingFreeStructuredDiffusion} or animals may have the wrong number of limbs, or appear as a head with no body (or vice-versa).  Because of these frequent failures, simply showing a randomly selected result has the risk of showing only a failure case,
whereas cherry picking a successful result is biased.
To address this we adopted the \SSatK protocol in which $k{=}12$ images are generated following from an initial random seed, and the best-of-k image is subjectively selected (cherry picked) for each method.
In this supplement however we show comparisons with \textbf{all 12} images and no subjective selection.


Please browse \FigRef{img:supp:church}--\FigRef{img:supp:multi:super} to examine our results relative to three concurrently developed competing methods.
The results of all four methods are broadly comparable. However our method sometimes performs better on specific issues such as color bleeding due to attribute binding failure (\FigRef{img:supp:church}), missing objects (\FigRef{img:supp:grave}), incorrect objects (\FigRef{img:supp:super}, \FigRef{img:supp:multi:super}), (im)plausible scene composition and layout (\FigRef{img:supp:iou}), object color (\FigRef{img:supp:dog}), and object-object interaction (\FigRef{img:supp:dog}).

Note that one of the most important factors in our paper is the interaction among the directed objects, other subjects that are not part of the algorithm, and the environment. Following the experiment section in the paper, we divide the experiments into \emph{Scene Compositing, \OneObject,} and \emph{\TwoObjects} tasks.

In the \textbf{Scene Compositing} task (Figs.~\ref{img:supp:church}--\ref{img:supp:cherry}), the prompt comprises an array of descriptive terms aimed at describing a holistic scene. For instance, consider the prompt \emph{``A white church under lightning in a pink sky.''} In this case, the attributes ``white'' and ``pink'' should be distinctly associated with the ``church'' and the ``sky'', respectively, to the greatest extent possible.

The \textbf{\OneObject} task (Figs.~\ref{img:supp:bonfire}--\ref{img:supp:grave}) explores how interactions manifest between a single directed object and other subjects not directed by the algorithm. For instance, when the directed object is ``bonfire'' and the prompt is ``A bonfire next to a man'', the synthesized image should not only depict the bonfire located at the specified region.
 but it should also portray the man positioned next to the bonfire, and with realistic interaction and lighting between the bonfire and the man.

The \OneObject task is extended to our final category, \textbf{\TwoObjects} (Figs.~\ref{img:supp:dog}--\ref{img:supp:multi:super}), which showcases control of \emph{two} directed objects as well as an optionally specified background.

In each experiment starting with \FigRef{img:supp:church}, 
we show the results of multiple random seeds for a single prompt,
allowing easy visual comparison of the quality and failures of each method. Bold text in the prompt indicates the directed object(s). Please enlarge the .pdf view to see details. The associated parameters and the commands in GLIGEN and BoxDiff are displayed below  each experiment.

\section{Quantitative Evaluation and Limitations}

In this section we discuss common quantitative metrics for generative models, and point out several reasons why these metrics are not as appropriate for our task as might be initially assumed. 

\textbf{CLIP score.} This metric evaluates whether the generated image successfully reflects the content of the prompt that was used. We report a quantitative evaluation using CLIP score \cite{CLIP} in the Experiments section of the main text. Although our results on this metric are favorable, we believe it is not the best measure for our task. Stable diffusion has considerable difficulty synthesizing scenes guided by a prompt containing several objects. This includes \emph{attribute binding} failures, in which a color or attribute that should be applied to one object is applied to others.
One hypothesis is that this difficulty with \SD is due to CLIP itself. Indeed, it appears that CLIP does not have a true grammatical understanding of the prompt \cite{TrainingFreeStructuredDiffusion,whatTheDAAM}
but rather behaves more as unordered sets of objects and attributes.  In this respect, while our method escapes these limitations of CLIP and allows more reliable synthesis of multiple objects, the CLIP score may be blind to the improvement that has been achieved.

\textbf{IOU.} IOU is very relevant for our position-directed generation task. In our results it can be seen that while the prominent portion of the synthesized object generally lies within the desired bounding box, some portions of it frequently extend outside the provided bounding box.
However, one subtlety in this task is that while position directed generative algorithms such as ours synthesize an object guided by a position indication such as a bounding box,  the position of the synthesized object is not explicitly given by the network itself.  One way of dealing with this is use a separate object segmentation method to attempt to locate the positioned object in the generated image.  An issue here is that there is no recognized baseline for this task and concurrent papers have selected different algorithms for this purpose.
The subjective choice of particular segmentation networks will lead to different IOU scores.

A larger issue is that achieving good IOU is somewhat in opposition to one of our goals (and a core strength of our method), specifically that of achieving strong \emph{interactions between objects}. For example, in the case of a prompt \emph{``a dog chasing a ball''}, a plausible interaction is that the dog is touching and partially obscured by the ball (see \FigRef{img:supp:dog}). This hinders the IOU since the estimated segmentation of the generated dog may not include the obscured region -- but it produces more natural interaction. To quantitatively illustrate this issue, in \FigRef{img:supp:iou} we show the IOUs of our synthesized objects together with those of two competing methods.  While our IOU results are comparable to one of the methods, and to some other results presented in the literature, they are clearly inferior to the other competing methods However the visual results of both our method
and the competing method with the lower IOU are \emph{better} than
the method with the higher IOU.
In other words, \emph{enforcing a high IOU appears to interfere with visual quality on our task.}

\textbf{FID.} The lack of relevance of FID for our task is direct: FID is intended to show improvements in generative models, however our innovation is not in the generative model but merely in how the objects are placed. Indeed, our method uses a pre-trained model with \emph{no modifications to the weights}.  Further, FID evaluates a distance between the distribution of representations of the new and baseline data in the representation layer of a classifier. Classifiers are designed to be \emph{position invariant}, e.g.~as implemented through pooling layers. In other words, FID is designed to be invariant to our task of changing the position of objects.

\section{Societal Impact}
The aim of this project is to extend the capability of text-to-image models to visual storytelling.
In common with most other technologies, there is potential for misuse. In particular,
generative models reflect the biases of their training data,
and there is a risk that malicious parties can use text-to-image methods to generate misleading images.
The authors believe that it is important that these risks be addressed. This might be done by penalizing malicious behavior
through the introduction of appropriate laws, or by limiting the capabilities of generative models to
serve this behavior.

\ifdefined\ARXIVMERGEDAPPENDIX
\else
  \bibliography{reference/references}
\fi

\onecolumn

\newpage
\subsection{Scene compositing: A white church under lightning in the pink sky}
\begin{figure}[H]
    \centering\captionsetup{skip=3pt}
    \includegraphics[width=.8\linewidth]{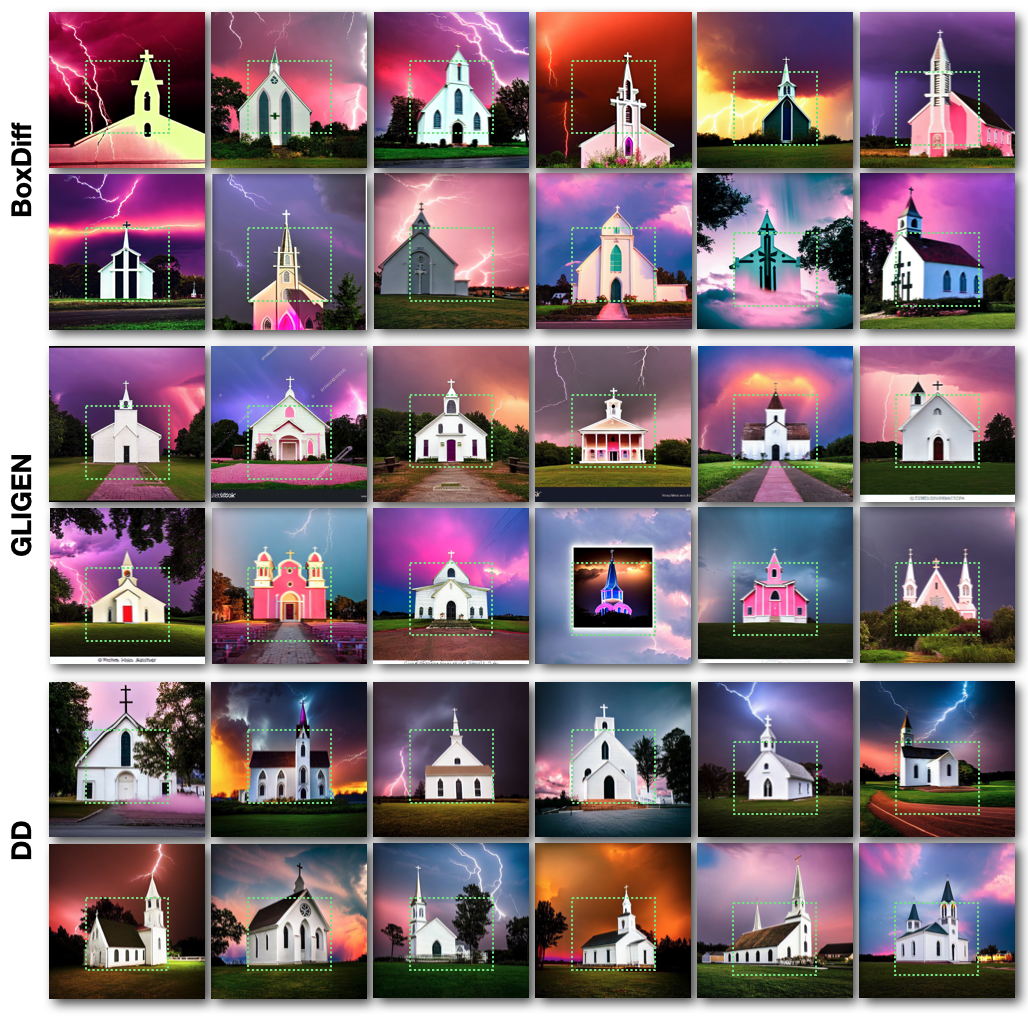}
    \caption{Prompt: A \textbf{white church} under lightning in the pink sky}
\label{img:supp:church}
\end{figure}
Some GLIGEN and BoxDiff results fall short in accurately disambiguating the ``pink'' attribute associated with the sky from the church. 
Additionally, certain outcomes from GLIGEN lack natural realism, featuring ``picture-in-picture'' scenarios (sometimes with captions) in which the church subject is unnaturally constrained to the position box. The DD results show a more natural variety of church orientations while still being placed in roughly the desired position.

The picture-in-picture behavior resembles our finding in \FigRef{img:trailing} (bottom row), in which excessively strong direction results in a nearly square burger in the desired bounding box with the squirrel and surrounding context missing.

\vspace{5mm}
The command and the code snippet for BoxDiff and GLIGEN:
\begin{small}
\begin{verbatim}
    # BoxDiff
    python run_sd_boxdiff.py --prompt "a white church under lightning in the pink sky" \
        --P 0.2 --L 1  --token_indices [3] --bbox [[82,131,357,383]]
    # GLIGEN
    dict(
        prompt = "a white church under lightning in the pink sky",
        phrases = ['a white church'],
        locations = [[0.25, 0.25, 0.75, 0.75]],
        alpha_type = [0.3, 0.0, 0.7],
    )
\end{verbatim}
\end{small}

\newpage
\subsection{Scene compositing: Mystical trees next to a dark magical pond}
\begin{figure}[H]
    \centering\captionsetup{skip=3pt}
    \includegraphics[width=.8\linewidth]{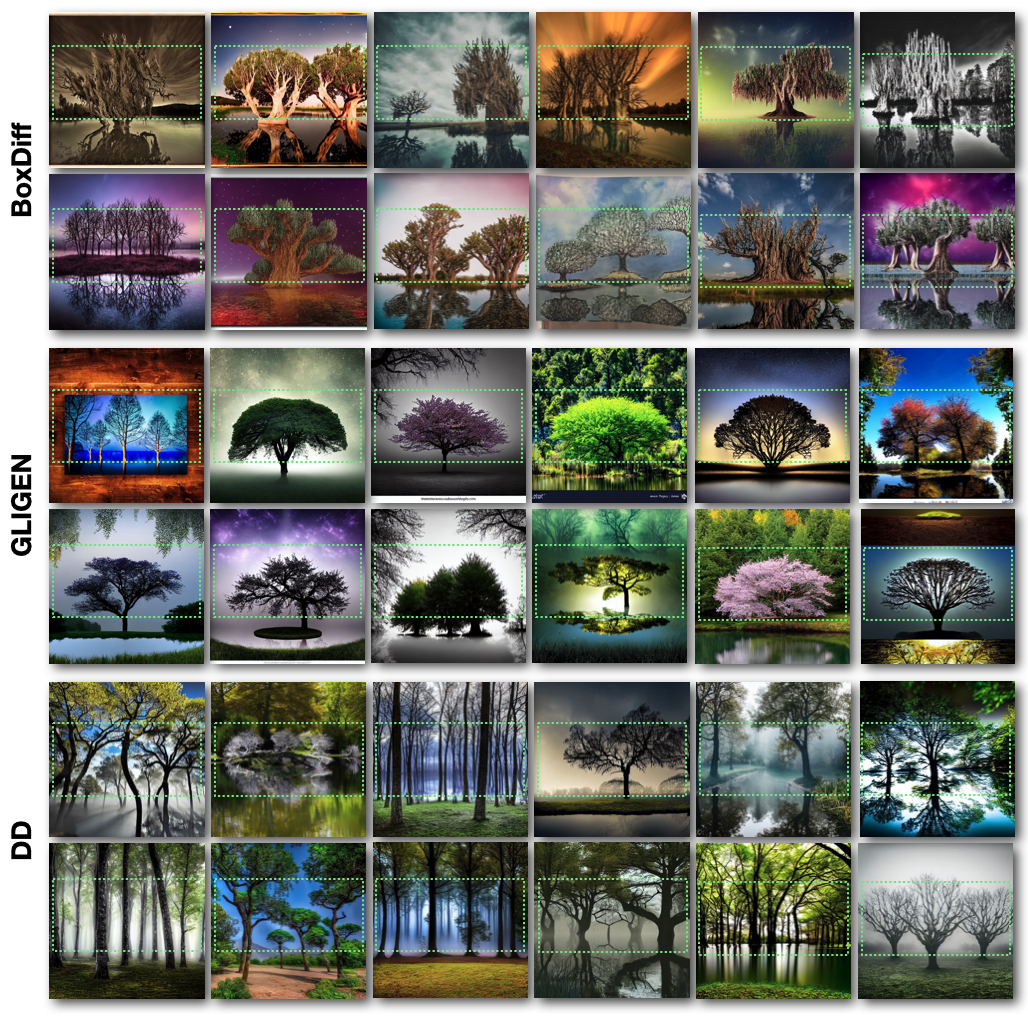}
    \caption{Prompt: \textbf{Mystical trees} next to a dark magical pond}
\label{img:supp:myth}
\end{figure}

The images produced by GLIGEN tend to position a single prominent tree within the bounding box, which contradicts the inclusion of multiple ```trees'' as specified in the prompt.

\vspace{5mm}
The command and the code snippet for BoxDiff and GLIGEN:
\begin{small}
\begin{verbatim}
    # BoxDiff
    python run_sd_boxdiff.py --prompt "mystical trees next to a dark magical pond" \
        --P 0.2 --L 1  --token_indices [2] --bbox [[46,86,504,385]]
    # GLIGEN
    dict(
        prompt = "mystical trees next to a dark magical pond",
        phrases = ['mystical trees'],
        locations = [[0.1, 0.3, 0.9, 0.8]],
        alpha_type = [0.3, 0.0, 0.7],
    )
\end{verbatim}
\end{small}

\newpage
\subsection{Scene compositing: A castle in a forest with grainy fog}
\begin{figure}[H]
    \centering\captionsetup{skip=3pt}
    \includegraphics[width=.8\linewidth]{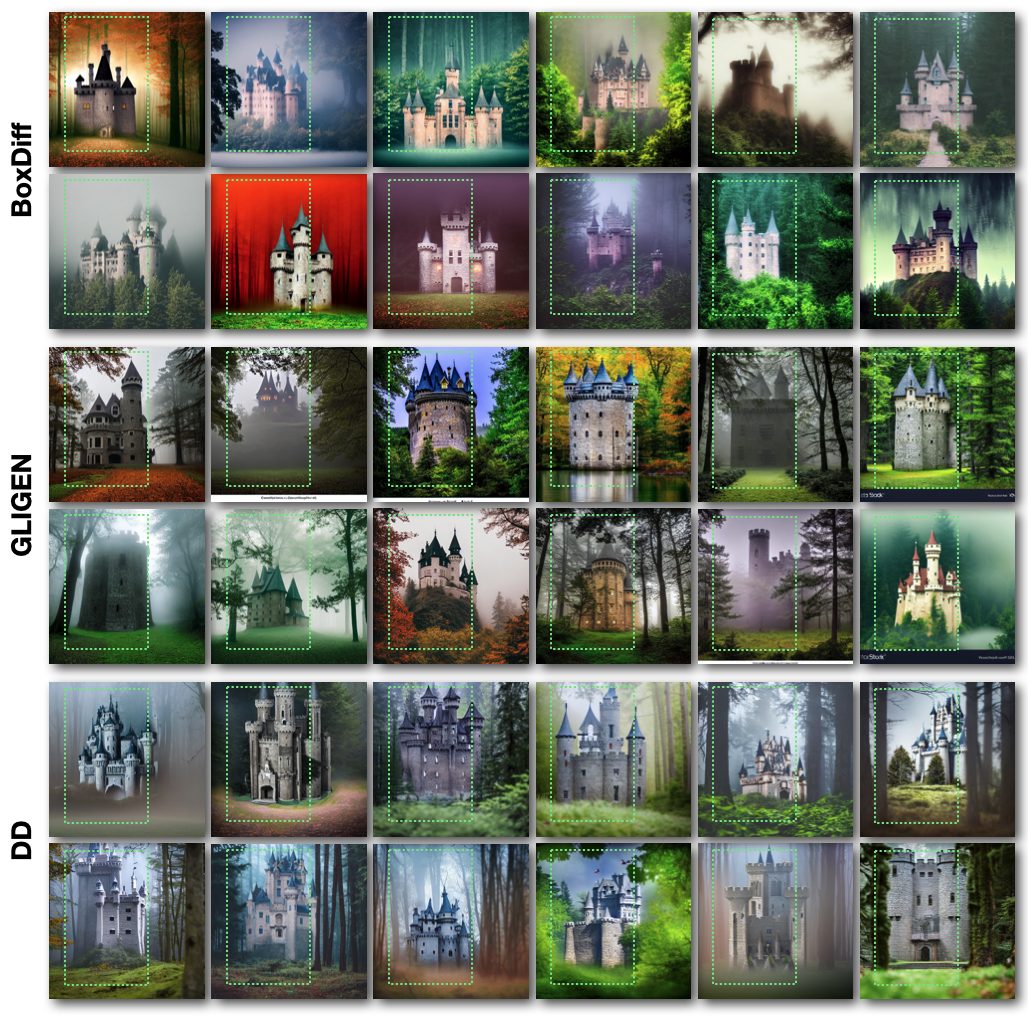}
    \caption{Prompt: A \textbf{castle} in a forest with grainy fog.}
\label{img:supp:castle}
\end{figure}

\vspace{5mm}
The command and the code snippet for BoxDiff and GLIGEN:
\begin{small}
\begin{verbatim}
    # BoxDiff
    python run_sd_boxdiff.py --prompt "a castle in a forest with grainy fog" \
        --P 0.2 --L 1  --token_indices [2] --bbox [[84,109,446,436]]
    # GLIGEN
    dict(
        prompt = "a castle in a forest with grainy fog",
        phrases = ['a castle'],
        locations = [[0.2, 0.1, 0.7, 0.8]],
        alpha_type = [0.3, 0.0, 0.7],
    )
\end{verbatim}
\end{small}

\newpage
\subsection{Scene compositing: cherry blossoms next to the lake and a mountain}
\begin{figure}[H]
    \centering\captionsetup{skip=3pt}
    \includegraphics[width=.8\linewidth]{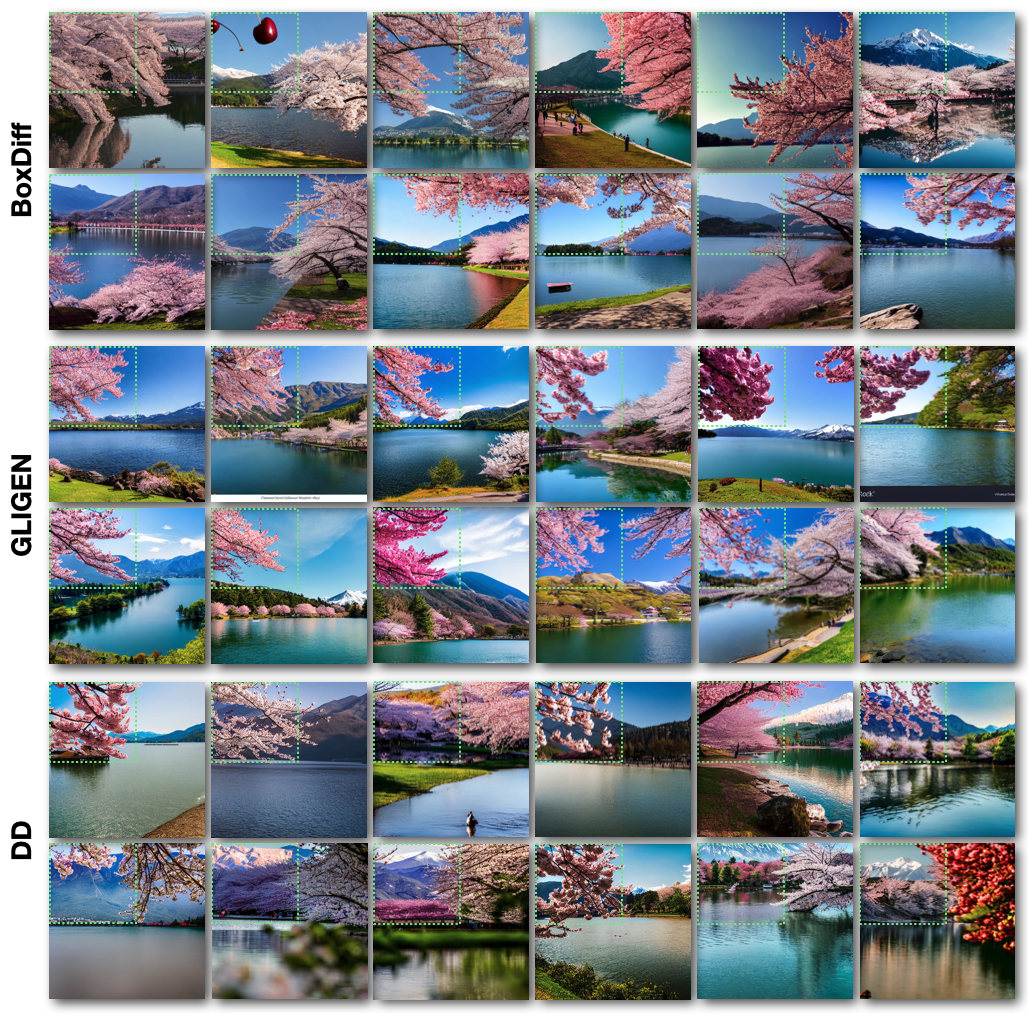}
    \caption{Prompt: The \textbf{cherry blossoms} next to the lake and a mountain}
\label{img:supp:cherry}
\end{figure}

\vspace{5mm}
The command and the code snippet for BoxDiff and GLIGEN:
\begin{small}
\begin{verbatim}
    # BoxDiff
    python run_sd_boxdiff.py --prompt "the cherry blossoms next to the lake and a mountain" \
        --P 0.2 --L 1  --token_indices [2] --bbox [[0,0,256,256]]
    # GLIGEN
    dict(
        prompt = "the cherry blossoms next to the lake and a mountain",
        phrases = ['the cherry blossoms'],
        locations = [[0.0, 0.0, 0.5, 0.5]],
        alpha_type = [0.3, 0.0, 0.7],
    )
\end{verbatim}
\end{small}

\newpage
\subsection{\OneObject: bonfire next to a man}
\begin{figure}[H]
    \centering\captionsetup{skip=3pt}
    \includegraphics[width=.8\linewidth]{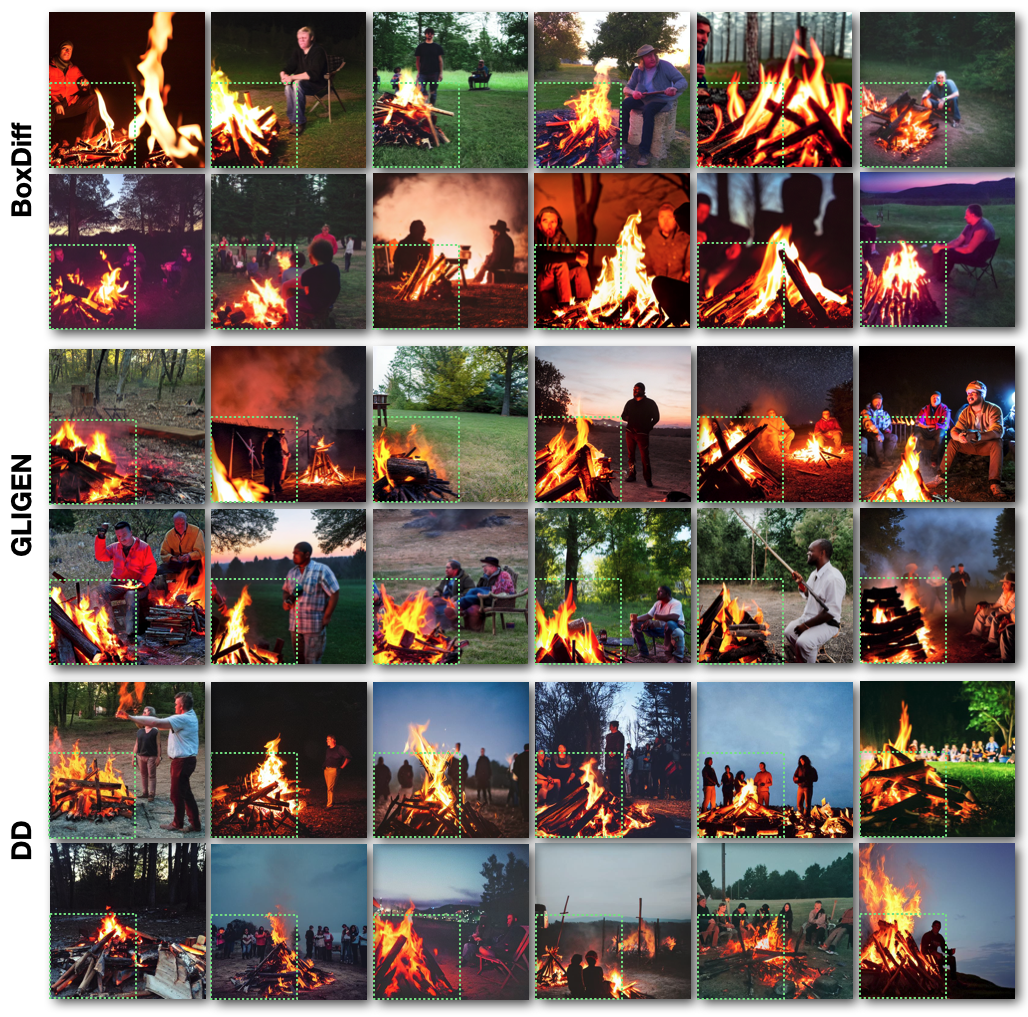}
    \caption{Prompt: The \textbf{bonfire} next to a man}
\label{img:supp:bonfire}
\end{figure}


\vspace{5mm}
The command and the code snippet for BoxDiff and GLIGEN:
\begin{small}
\begin{verbatim}
    # BoxDiff
    python run_sd_boxdiff.py --prompt "the bonfire next to a man" \
        --P 0.2 --L 1  --token_indices [2] --bbox [[0,256,256,512]]
    # GLIGEN
    dict(
        prompt = "the bonfire next to a man",
        phrases = ['the bonfire'],
        locations = [[0.0, 0.5, 0.5, 1.0]],
        alpha_type = [0.3, 0.0, 0.7],
    )
\end{verbatim}
\end{small}

\newpage
\subsection{\OneObject: A white horse in front of the erupting volcano}
\begin{figure}[H]
    \centering\captionsetup{skip=3pt}
    \includegraphics[width=.8\linewidth]{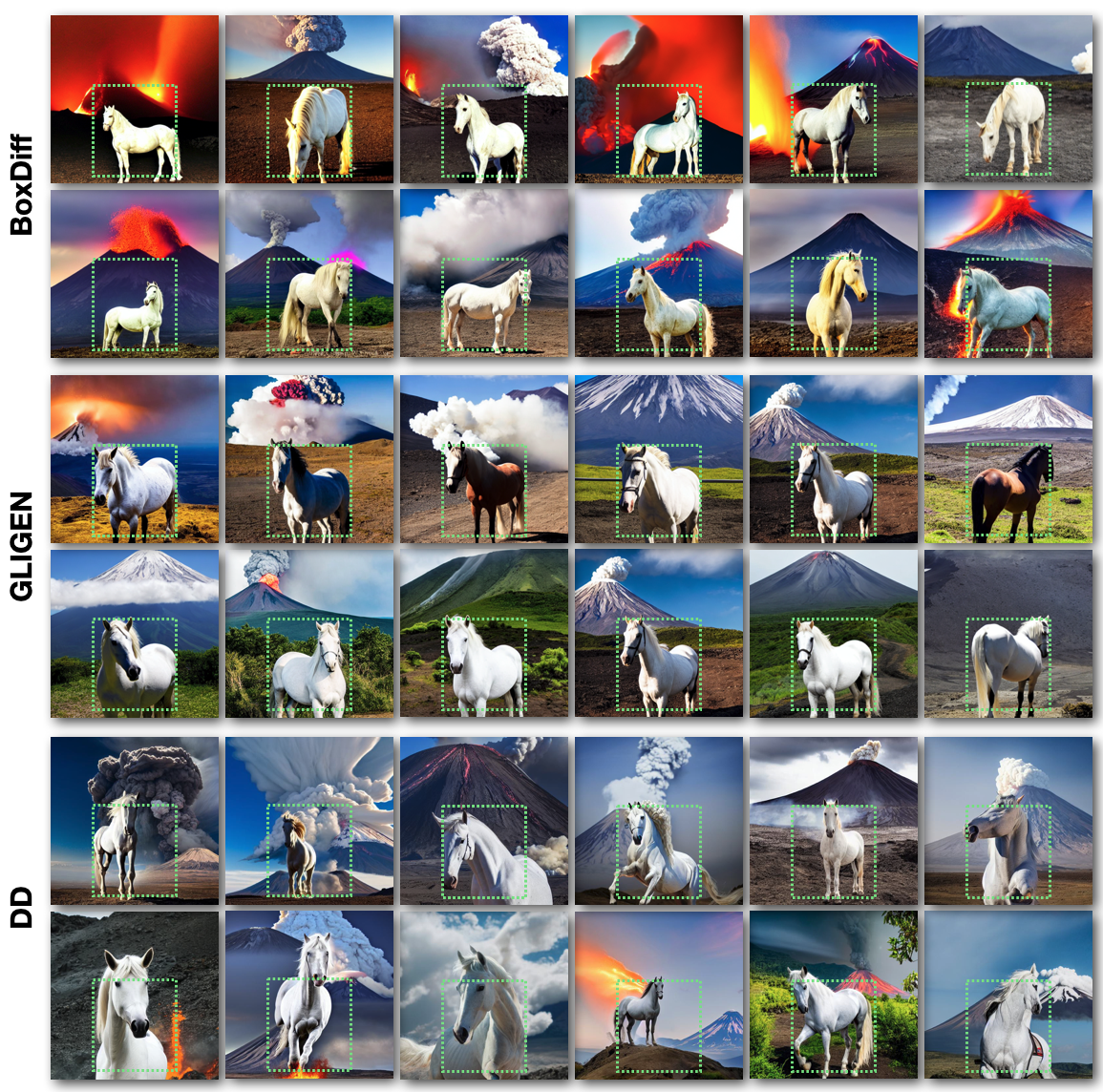}
    \caption{Prompt: A \textbf{white horse} in front of the erupting volcano}
\label{img:supp:horse}
\end{figure}

\vspace{5mm}
The command and the code snippet for BoxDiff and GLIGEN:
\begin{small}
\begin{verbatim}
    # BoxDiff
    python run_sd_boxdiff.py --prompt "a white horse in front of the erupting volcano" \
        --P 0.2 --L 1  --token_indices [3] --bbox [[150,235,423,494]]
    # GLIGEN
    dict(
        prompt = "a white horse in front of the erupting vocano",
        phrases = ['white horse'],
        locations = [[0.25, 0.4, 0.75, 1.0]],
        alpha_type = [0.3, 0.0, 0.7],
    )
\end{verbatim}
\end{small}

\newpage
\subsection{\OneObject: A gravestone next to a man in red shirt}
\begin{figure}[H]
    \centering\captionsetup{skip=3pt}
    \includegraphics[width=.8\linewidth]{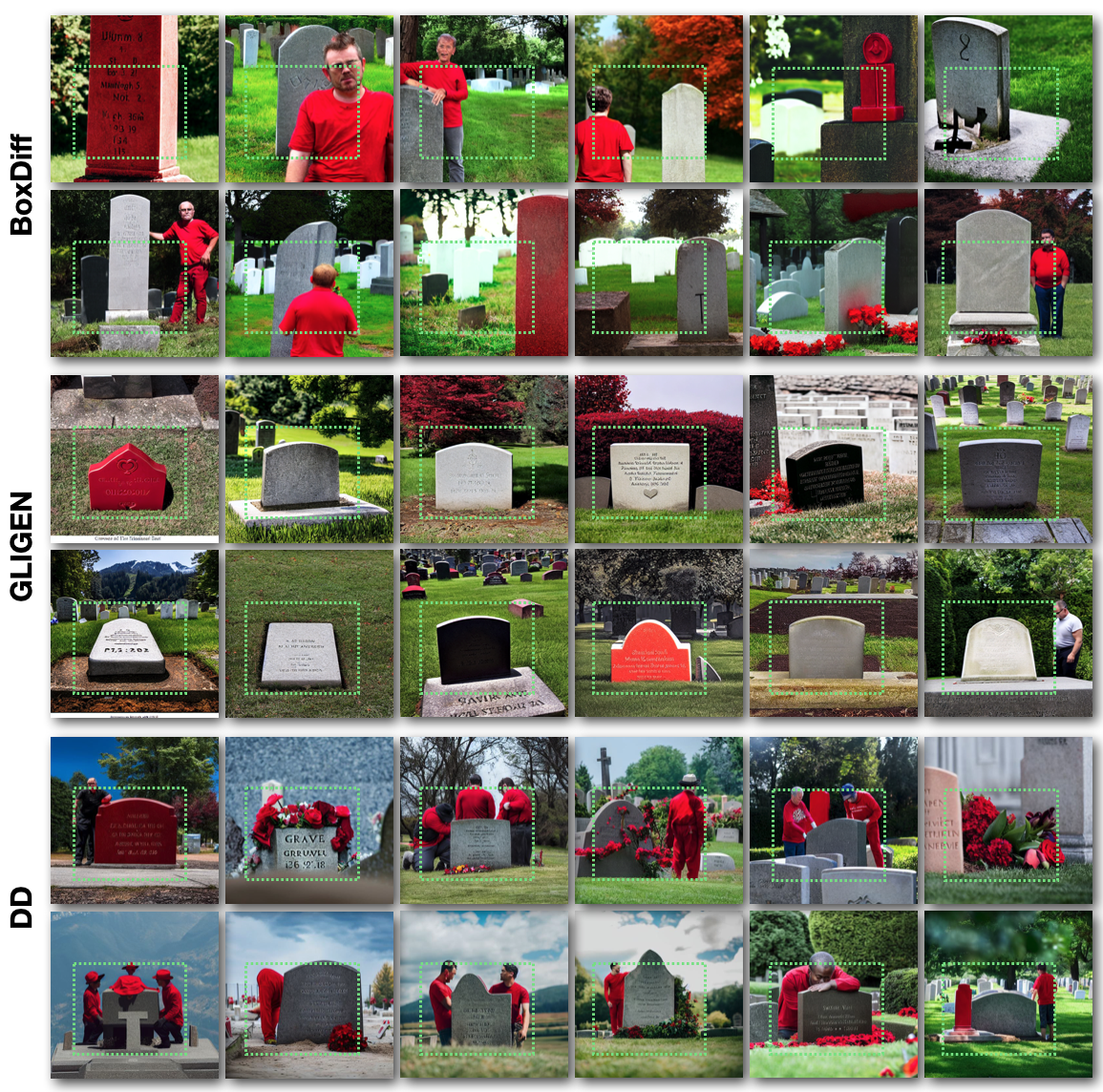}
    \caption{Prompt: A \textbf{gravestone} next to a man in red shirt}
\label{img:supp:grave}
\end{figure}

The outcomes from GLIGEN do not align with the prompt (e.g., a man in a red shirt), whereas the directed object ``gravestone'' consistently fits within the bounding box across all instances. Our method sometimes fails to generate the man or generates multiple men, although several seeds are successful.

\vspace{5mm}
The command and the code snippet for BoxDiff and GLIGEN:
\begin{small}
\begin{verbatim}
    # BoxDiff
    python run_sd_boxdiff.py --prompt "a gravestone next to a man in red shirt" \
        --P 0.2 --L 1  --token_indices [2] --bbox [[48,165,461,438]]
    # GLIGEN
    dict(
        prompt = "a gravestone next to a man in red shirt",
        phrases = ['gravestone'],
        locations = [[0.2, 0.4, 0.7, 0.8]],
        alpha_type = [0.3, 0.0, 0.7],
    )
\end{verbatim}
\end{small}



\newpage
\subsection{\TwoObjects: A white dog chasing a red sphere}
\begin{figure}[H]
    \centering\captionsetup{skip=3pt}
    \includegraphics[width=.8\linewidth]{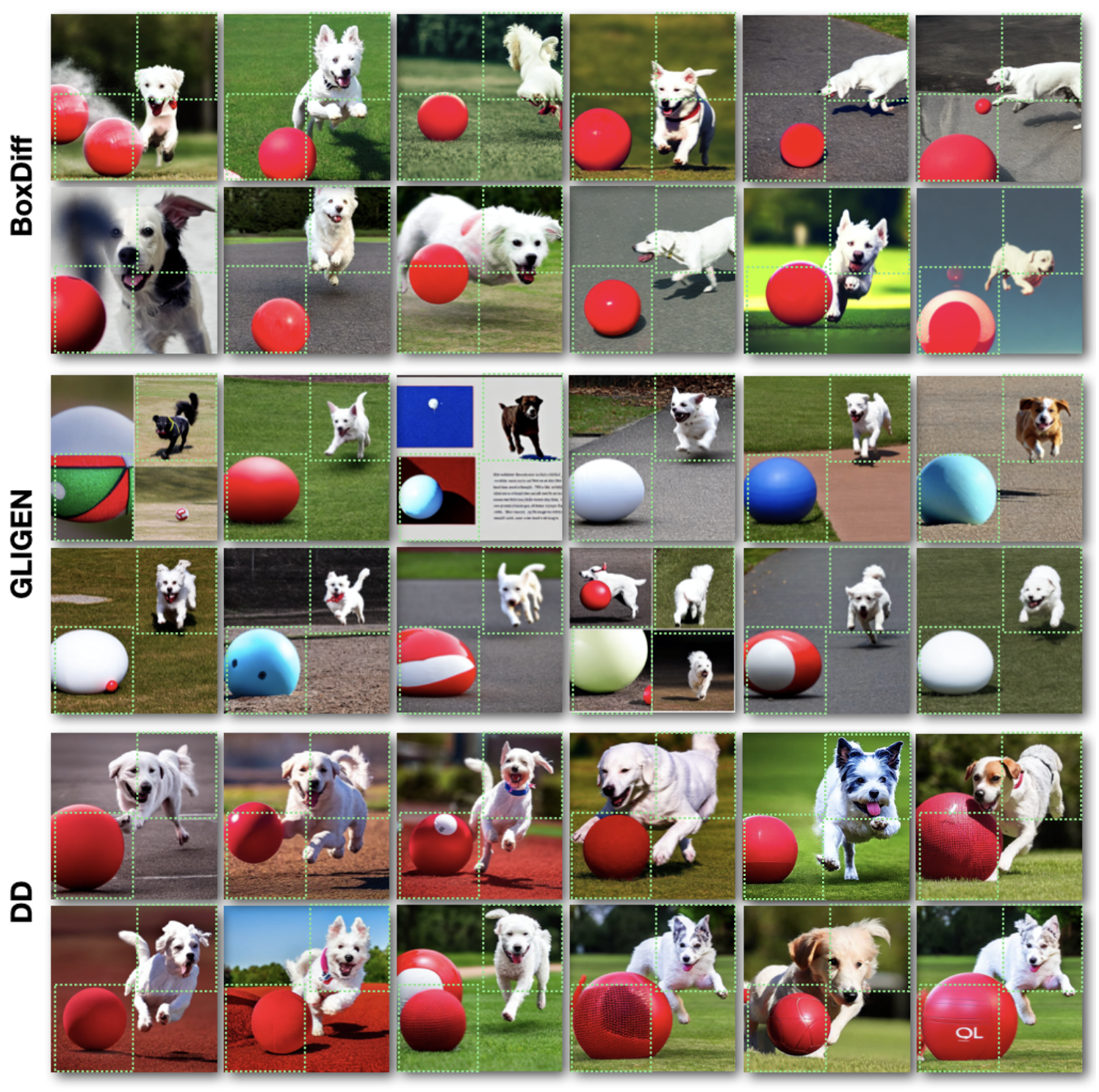}
    \caption{Prompt: A \textbf{white dog} chasing a red sphere.}
\label{img:supp:dog}
\end{figure}
A majority of the outcomes produced by GLIGEN are flawed (e.g., generating four subfigures inside the overall image or depicting the dog and sphere on a page of a book) or exhibit incorrect coloring for the ball and dog. The BoxDiff results are more successful with only a couple failures (missing background, dog running away from the sphere). The DD results are the most natural, although there is leakage of the red color to the background in a couple cases.

\vspace{5mm}
The command and the code snippet for BoxDiff and GLIGEN:
\begin{small}
\begin{verbatim}
    # BoxDiff
    python run_sd_boxdiff.py --prompt "a white dog chasing a red ball" \
        --P 0.2 --L 1 --token_indices [3,7] --bbox [[243,6,506,263],[5,265,244,501]]
    # GLIGEN
    dict(
        prompt = "A white running dog chasing a red ball",
        phrases = ['white running dog', 'red ball'],
        locations = [(0.6, 0.1, 0.9, 0.5), (0.0, 0.5, 0.5, 0.9)],
        alpha_type = [0.3, 0.0, 0.7],
    )
\end{verbatim}
\end{small}

\newpage
\subsection{\TwoObjects: A red cube above the blue sphere in the supermarket}
\begin{figure}[H]
    \centering\captionsetup{skip=3pt}
    \includegraphics[width=.8\linewidth]{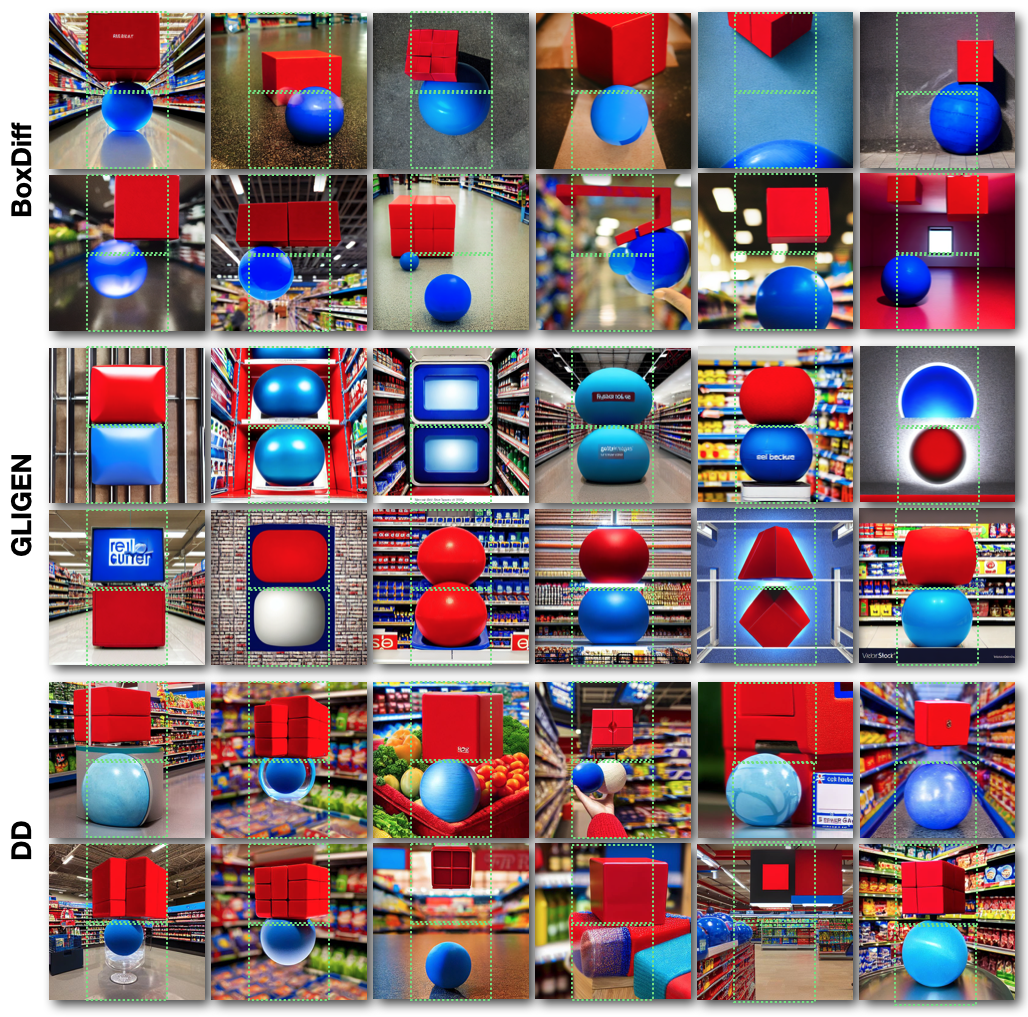}
    \caption{Prompt: A \textbf{red cube} above the \textbf{blue sphere} in the supermarket.}
\label{img:supp:super}
\end{figure}
The outcomes obtained from GLIGEN generally do not faithfully replicate the (cube, sphere) objects requested in the prompt.
The BoxDiff results correctly show a red cube above a blue sphere,
but fail to show a supermarket background in some cases.
Our results have the best alignment to the prompt, including a nice example where the red cube and blue sphere are placed in a basket brimming with vegetables and fruits.

\vspace{5mm}
The command and the code snippet for BoxDiff and GLIGEN:
\begin{small}
\begin{verbatim}
    # BoxDiff
    python run_sd_boxdiff.py --prompt "a red cube above the blue sphere in the supermarket" \
        --P 0.2 --L 1 --token_indices [3,7] --bbox [[93,4,447,232],[93,244,452,504]]
    # GLIGEN
    dict(
        prompt = "a red cube above the blue sphere in the supermarket",
        phrases = ['a red cube', 'a blue sphere'],
        locations = [ [0.25,0.1,0.75,0.5], [0.25,0.5,0.75,0.9] ],
        alpha_type = [0.3, 0.0, 0.7],
    )
\end{verbatim}
\end{small}

\newpage
\subsection{\TwoObjects: MultiDiffusion}

\begin{figure}[H]
    \centering\captionsetup{skip=1pt}
    \includegraphics[width=.7\linewidth]{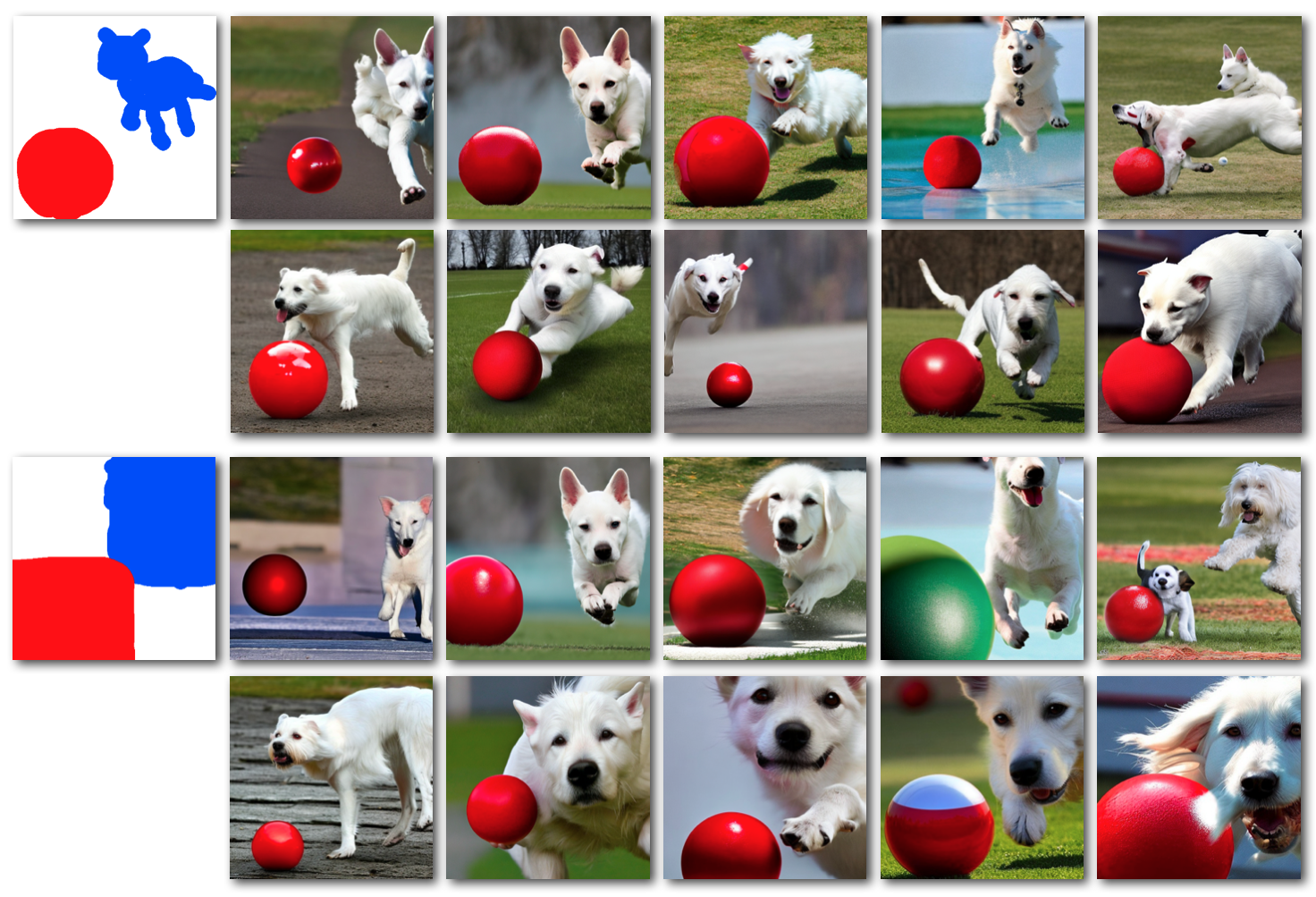}
    \caption{Prompt: A \textbf{white dog} chasing after the \textbf{red ball}}
\label{img:supp:multi:dog}
\end{figure}

\begin{figure}[H]
    \centering\captionsetup{skip=1pt}
    \includegraphics[width=.7\linewidth]{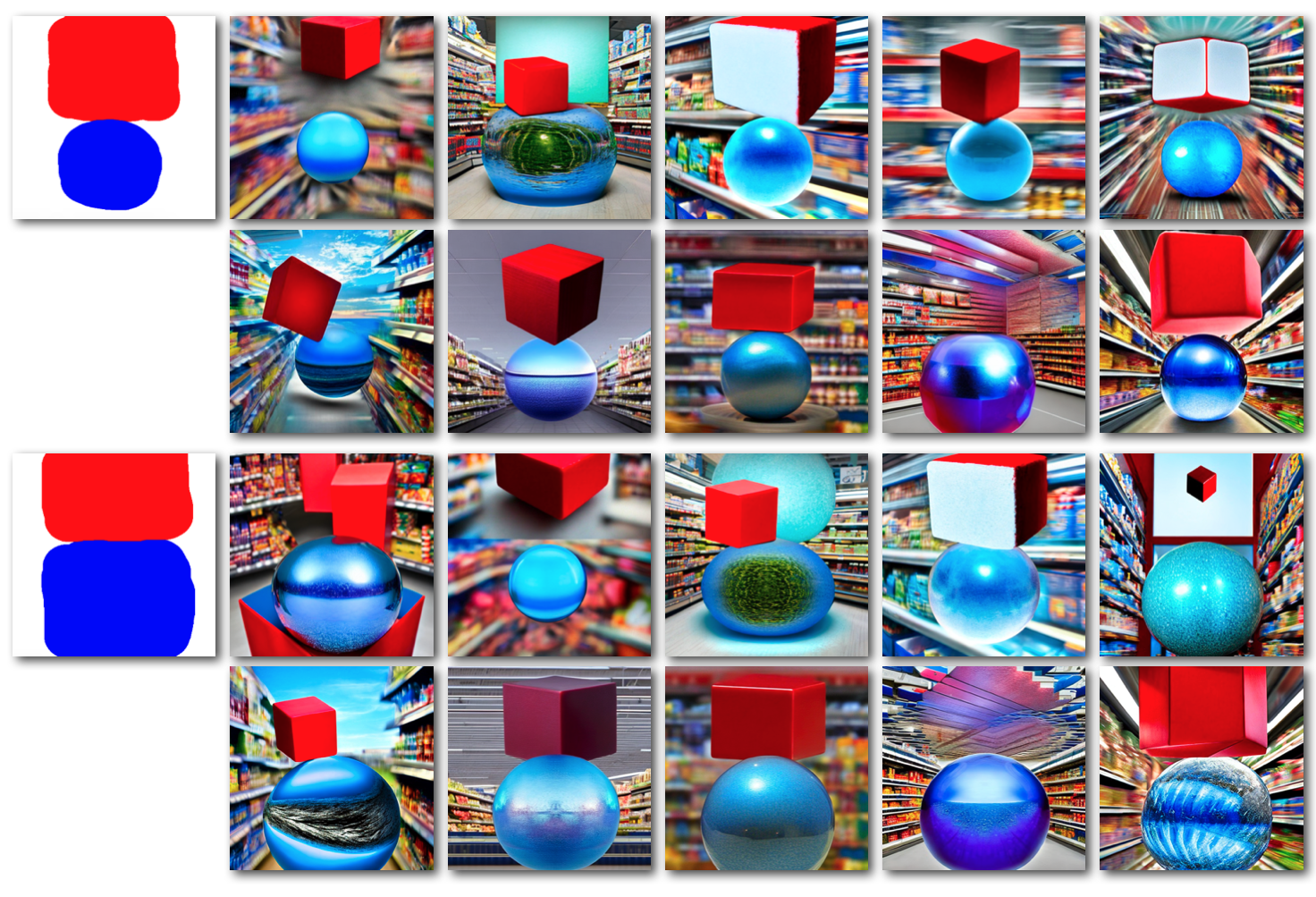}
    \caption{Prompt: A \textbf{red cube} above the \textbf{blue sphere} in the supermarket}
\label{img:supp:multi:super}
\end{figure}

Here we present the outcome from MultiDiffusion \cite{multidiffusion}. \FigRef{img:supp:multi:dog} and \FigRef{img:supp:multi:super} replicate the experiments previously conducted in \FigRef{img:supp:dog} and \FigRef{img:supp:super}. MultiDiffusion uses masks rather than boxes as its position guidance, so we provided both shaped masks and (approximate) boxes to provide a fair comparison.
The method does not appear to closely follow the detailed mask, however.
In their interface the color of the mask denotes a particular object but does not reflect the requested color of the object.
The default demo parameters are used, with only the random seed changing.
Please refer to the MultiDiffusion public Huggingface demo for more detail. \footnote{https://huggingface.co/spaces/weizmannscience/multidiffusion-region-based}

\newpage
\subsection{Intersection over Union}
\begin{figure}[H]
    \centering\captionsetup{skip=3pt}
    \includegraphics[width=.8\linewidth]{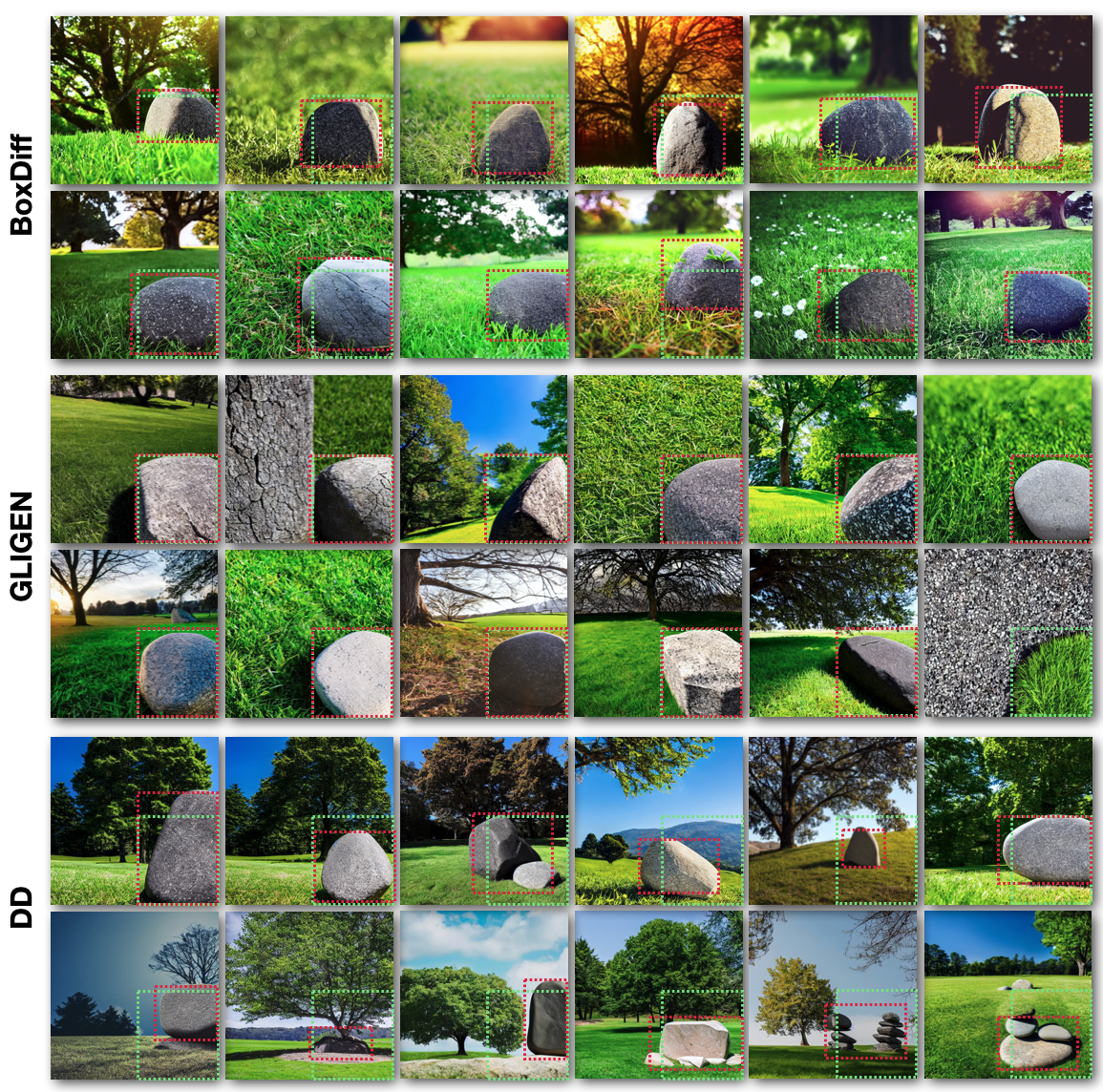}
    \caption{Prompt: The \textbf{stone} on grass field in sunny day next to a tree}
\label{img:supp:iou}
\end{figure}

In the \FigRef{img:supp:iou}, we present Intersection over Union (IOU) results of the experiment depicted in \FigRef{img:mask}. The green bounding box is the guidance (groundtruth) given to the algorithms, and the red bounding box is an estimate of the location of the synthesized object; this was produced using a simple box selection user interface by a person not associated with the project.

The averaged IOU values for BoxDiff, GLIGEN, and our DD approach are 0.51, 0.87, and \textbf{0.53}, respectively. Notably, the average IOU for GLIGEN is nearly twice as high as those for BoxDiff and DD. As seen in this example, it is important to emphasize that the IOU metric \emph{does not correlate to the quality of the results} since our objective is not only to control the object placement but \emph{also} generate natural images with plausible composition and interaction of the object and environment.

The command and the code snippet for BoxDiff and GLIGEN:
\begin{small}
\begin{verbatim}
    # BoxDiff
    python run_sd_boxdiff.py --prompt "the stone on grass field in sunny day next to a tree" \
        --P 0.2 --L 1  --token_indices [2] --bbox [[256,256,512,512]]
    # GLIGEN
    dict(
        prompt = "the stone on grass field in sunny day next to a tree",
        phrases = ['stone'],
        locations = [[0.5, 0.5, 1.0, 1.0]],
        alpha_type = [0.3, 0.0, 0.7],)
\end{verbatim}
\end{small}

\ifdefined\AAAISUBMISSIONONLY
\newpage
\setlength{\parindent}{0ex}
\setlength{\parskip}{1.5em}
\setlength{\itemsep}{0.5em}

\section{Reproducability Checklist}
\textbf{This paper:}
\begin{itemize}
\item \textbf{Includes a conceptual outline and/or pseudocode description of AI methods introduced?} yes: Pseudocode in Algo.~1.

\item \textbf{Clearly delineates statements that are opinions, hypothesis, and speculation from objective facts and results?} yes.

\item \textbf{Provides well marked pedagogical references for less-familiar  readers to gain background necessary to replicate the paper?} Yes. The related work section cites the excellent tutorial by L.~Weng.
\end{itemize}

\textbf{Does this paper make theoretical contributions?} No.\\
The paper introduces a new algorithm, but this is an applied paper. There are no theorems.

\textbf{Does this paper rely on one or more datasets?} No.\\
We use the pretrained Stable Diffusion model \cite{RombachStablediffusion21}. No datasets or training is involved.

\textbf{Does this paper include computational experiments?} yes.

\begin{itemize}
\item \textbf{Any code required for pre-processing data is included in the appendix?} N/A. No data is needed.

\item \textbf{All source code required for conducting and analyzing the experiments is included in a code appendix?} N/A.\\
Full source code will be made public upon publication of the paper.

\item \textbf{All source code required for conducting and analyzing the experiments will be made publicly available upon publication of the paper with a license that allows free usage for research purposes?} Yes.

\item \textbf{All source code implementing new methods have comments detailing the implementation, with references to the paper where each step comes from?} Yes.

\item \textbf{If an algorithm depends on randomness, then the method used for setting seeds is described in a way sufficient to allow replication of results.} Yes. See the Qualitative Experiments section in the supplementary, and (in shorter form) in the Experiments and Comparisons section in the main paper.

\item \textbf{This paper specifies the computing infrastructure used for running experiments (hardware and software), including GPU/CPU models; amount of memory; operating system; names and versions of relevant software libraries and frameworks.} Yes, in the supplementary section.

\item \textbf{This paper formally describes evaluation metrics used and explains the motivation for choosing these metrics.} Yes. The supplementary includes a section discussing appropriate metrics as well as some limitations.

\item \textbf{This paper states the number of algorithm runs used to compute each reported result.} Yes (or N/A). There is only one algorithm run per seed.

\item \textbf{Analysis of experiments goes beyond single-dimensional summaries of performance (e.g., average; median) to include measures of variation, confidence, or other distributional information.} No (N/A).
There is no variation other than the random seed.

\item \textbf{The significance of any improvement or decrease in performance is judged using appropriate statistical tests (e.g., Wilcoxon signed-rank).} No. N/A

\item \textbf{This paper lists all final (hyper-)parameters used for each model/algorithm in the paper’s experiments.} Yes.

\item \textbf{This paper states the number and range of values tried per (hyper-) parameter during development of the paper, along with the criterion used for selecting the final parameter setting.} Partial. (See ablation section in the supplementary).

\end{itemize}

\fi

\end{document}